\documentclass[11pt]{article}

\usepackage[preprint]{acl}

\usepackage{times}
\usepackage{latexsym}

\usepackage[T1]{fontenc}

\usepackage[utf8]{inputenc}

\usepackage{microtype}

\usepackage{inconsolata}

\usepackage{graphicx}

\usepackage{hyperref}
\usepackage{url}
\usepackage{booktabs}
\usepackage{array}
\usepackage{xcolor}
\usepackage{enumitem}
\usepackage{multirow}
\usepackage{subcaption}
\usepackage{float}
\usepackage{adjustbox}
\usepackage{tikz}
\usetikzlibrary{positioning,calc,fit,backgrounds,arrows.meta}
\usepackage{pgfplots}
\pgfplotsset{compat=1.16}
\usepackage{pifont}
\newcommand{\cmark}{\ding{51}}
\newcommand{\xmark}{\ding{55}}
\usepackage{listings}
\definecolor{okgreen}{HTML}{4f8a5b}
\definecolor{badred}{HTML}{b5523b}
\definecolor{slate}{HTML}{3b5b8c}
\definecolor{amber}{HTML}{c08a2d}

\pgfplotsset{
  paperbars/.style={ybar, clip=false, ymin=0, ymax=1.1, axis lines*=left, ymajorgrids, grid style={black!10},
    tick label style={font=\scriptsize}, label style={font=\scriptsize}, ylabel style={yshift=-4pt},
    ytick={0,0.25,0.5,0.75,1}, yticklabels={0,.25,.50,.75,1}, xtick style={draw=none},
    xticklabel style={font=\scriptsize, align=center},
    error bars/y dir=both, error bars/y explicit, error bars/error bar style={line width=0.4pt, black!70},
    error bars/error mark options={rotate=90, mark size=1.2pt, black!70}},
  armnone/.style={fill=black!35, draw=none},
  armvalid/.style={fill=okgreen, draw=none},
  armood/.style={fill=badred, draw=none}}
\newcommand{\armkey}[1]{\tikz[baseline=-0.6ex]\fill[#1] (0,-0.08) rectangle (0.2,0.12);}
\newcommand{\armlegend}{{\scriptsize\armkey{black!35}~No experience\quad\armkey{okgreen}~Valid experience\quad
  \armkey{badred}~Out-of-date experience}}
\newcommand{\code}[1]{{\upshape\ttfamily #1}}
\title{Experience Rot:\\Experienced Agents Fail Where Novices Succeed}

\author{
    Zhen Xu \\
    University of Chicago \\
    \And
    Hang Yin \\
    eBay \\
    \And
    Zitao Zhang \\
    eBay \\
    \And
    Kevin Zhao \\
    eBay \\
    \And
    Xuefeng Liu \\
    University of Florida
}

\begin{document}
\maketitle

\begin{abstract}
Agents are increasingly built to learn from experience: they keep conversation histories, memory files, retrieved notes
and skills, and reuse what worked before. We show that this experience turns against them when their tools change
silently. We test a range of models with tools defined as in the Model Context Protocol (MCP). After a silent
change, experienced agents fail on most tasks, while agents new to the task mostly succeed. We call this failure
Experience Rot. Experienced agents copy their earlier call instead of checking the current interface, even when
the tool definition shows the change. Then they report success. When the documentation is incomplete, experience is what makes agents succeed. After a change, the same experience leaves them worse off than no experience
at all. The failure appears in every form of memory we test and grows with the amount of experience. Simple remedies
fall short: announcing the update barely helps, and pointing the agent to the documentation fails whenever the
documentation does not spell out the answer.
\end{abstract}

\section{Introduction}
\label{sec:intro}

\begin{figure*}[t]
\centering
% Figure 1(a): deployed interfaces. Every entry is quoted from the cited source (see the caption in introduction.tex).
\begin{tikzpicture}[font=\scriptsize,
  card/.style={inner sep=3.5pt, anchor=north west, text width=0.32\textwidth-7pt, align=left}]
\node[anchor=south west, inner sep=0pt, font=\footnotesize\bfseries] at (0,0.12) {(a) In deployed interfaces, the
same field can mean opposite things};
\node[card] (c1) at (0,0) {%
{\bfseries\color{slate} One integer, two directions}\\[1pt]
Todoist task \code{priority}, an integer from 1 to 4.\\
Which value is the highest priority?\\[1.5pt]
\setlength{\tabcolsep}{0pt}%
\begin{tabular}{@{}p{\dimexpr\linewidth-12pt\relax}r@{}}
API, sync command \emph{add task} & \textbf{\color{badred}4}\\
API, endpoint \emph{create task} & \textbf{\color{slate}1}\\
MCP server A & \textbf{\color{slate}1}\\
MCP server B & \textbf{\color{badred}4}\\
\end{tabular}};
\node[card] (c2) at ($(c1.north east)+(0.0175\textwidth,0)$) {%
{\bfseries\color{slate} One integer, two weekdays}\\[1pt]
Day of the week, an integer from 0 to 6.\\
Which day is 0?\\[1.5pt]
\setlength{\tabcolsep}{0pt}%
\begin{tabular}{@{}p{\dimexpr\linewidth-36pt\relax}r@{}}
Python, \code{date.weekday()} & \textbf{\color{slate}Monday}\\
JavaScript, \code{Date.getDay()} & \textbf{\color{badred}Sunday}\\
POSIX, \code{crontab} & \textbf{\color{badred}Sunday}\\
\end{tabular}};
\node[card] (c3) at ($(c2.north east)+(0.0175\textwidth,0)$) {%
{\bfseries\color{slate} Changes need not show}\\[1pt]
An MCP server can replace its tool definitions during a session
(\code{notifications/tools/list\_changed}); notes from earlier sessions stay as they were.\\[2pt]
Four of five production API endpoints tested by SilentProbe ignore an undeclared parameter without an error.};
% equal-height card frames behind the three text nodes
\path let \p1=(c1.south), \p2=(c2.south), \p3=(c3.south), \n1={min(\y1,\y2,\y3)} in coordinate (bot) at (0,\n1);
\begin{scope}[on background layer]
\foreach \c in {c1,c2,c3}
  \draw[draw=black!18, fill=black!2.5, rounded corners=2pt] (\c.north west) rectangle (\c.east |- bot);
\end{scope}
\end{tikzpicture}\\[3.5mm]
\begin{minipage}[t]{0.66\textwidth}
\vspace{0pt}
% Figure 1(b): two runs of DeepSeek-V4-Flash on the same final task (rot6-v1, Experiment 1, change visible in the tool
% definition, field name attendees -> guests, direction A to B, repeat 2): no experience, and out-of-date experience.
% Calls and replies are quoted from the run records and shortened where marked with ... or \dots.
\begin{tikzpicture}[font=\scriptsize]
\pgfmathsetlengthmacro{\W}{\linewidth}
\pgfmathsetlengthmacro{\cA}{0.355*\W}   % earlier-task column
\pgfmathsetlengthmacro{\xF}{0.415*\W}   % left edge of the final-task column
\pgfmathsetlengthmacro{\cF}{0.585*\W}   % final-task column
\tikzset{
  box/.style={rounded corners=2pt, inner sep=3pt, anchor=north west, align=left},
  colhead/.style={font=\scriptsize\bfseries, text=black!60, anchor=north west, inner sep=0pt}}
\node[anchor=south west, inner sep=0pt, font=\footnotesize\bfseries] at (0,0.12) {(b) The same final task, without and
with experience};
% shared final request and tool definition, given to both agents
\node[box, fill=slate!8, text width=\W-7pt] (req) at (0,0) {%
{\bfseries\color{slate} Final task on version B, the same for both agents}\\[1pt]
\textit{User:} Schedule a 60-minute event titled ``Security Review'' on March 25, 2026 at 1:00 PM \dots{} with Grace
and Erin as participants.\\
\textit{Tool definition:} \code{calendar\_create\_event(payload: \{title, start\_time, duration\_minutes,
\textbf{guests}\})}};
\node[colhead] (h3) at ($(req.south west)+(\xF,-0.16)$) {Response to the final task};
% row 1: no experience
\node[box, draw=okgreen!80!black, fill=okgreen!6, text width=\cF-7pt] (f1) at ($(h3.south west)-(0,0.1)$) {%
\code{calendar\_create\_event(\{..., "guests": [...]\})}\\
$\rightarrow$ \code{\{"accepted": true\}}\\
{\color{okgreen!80!black}\cmark\ Stored event: Grace and Erin take part.}};
\node[anchor=north west, inner sep=0pt, align=left, text width=\cA] at (0,0 |- f1.north) {%
{\bfseries\color{okgreen!80!black} Without experience}\\[1pt]{\color{black!45}\itshape no earlier task}};
% row 2: out-of-date experience
\node[box, text width=\cF-7pt] (f2) at ($(f1.south west)-(0,0.16)$) {%
\code{calendar\_create\_event(\{..., "attendees": [...]\})}\\
$\rightarrow$ \code{\{"accepted": true\}}\\
\textit{Agent:} I've scheduled the event successfully \dots{} Attendees: grace@example.com, erin@example.com\\
{\color{badred}\xmark\ Stored event: no participants.}};
\node[box, text width=\cA-7pt] (e2) at (0,0 |- f2.north) {%
{\bfseries\color{badred} With experience}\\
{\bfseries Earlier task, version A}\\[1pt]
\code{calendar\_create\_event(}\\
\code{\ \ \{..., "attendees": [...]\})}\\
$\rightarrow$ \code{\{"accepted": true\}}\\
{\color{okgreen!80!black}\cmark\ Stored: Frank and Paul.}};
% frames of the second row share one bottom edge
\path let \p1=(e2.south), \p2=(f2.south), \n1={min(\y1,\y2)} in coordinate (bot2) at (0,\n1);
\begin{scope}[on background layer]
\draw[rounded corners=2pt, draw=black!30, fill=white] (e2.north west) rectangle (e2.east |- bot2);
\draw[rounded corners=2pt, draw=badred, fill=badred!5] (f2.north west) rectangle (f2.east |- bot2);
\end{scope}
\draw[-{Stealth[length=4pt]}, line width=0.9pt, badred] (e2.east) -- (e2.east -| f2.west)
  node[midway, above=0.5pt, font=\tiny\itshape, text=badred, align=center, inner sep=0pt] {silent\\change};
\end{tikzpicture}
\end{minipage}\hfill
\begin{minipage}[t]{0.32\textwidth}
\vspace{0pt}
{\footnotesize\bfseries (c) Success after a silent change}\\[1mm]
% Generated by prior_drift/paper_analysis/rot_figures.py (Experiment 6: ambig6-v1, told arms). Do not edit by hand.
\begin{tikzpicture}
\begin{axis}[ybar, width=0.95\linewidth, height=4.6cm, bar width=20pt, ymin=0, ymax=1.12,
  ytick={0,0.25,0.5,0.75,1}, yticklabels={0,.25,.50,.75,1}, ylabel={Success on the final task},
  ylabel style={font=\scriptsize, yshift=-3pt}, tick label style={font=\scriptsize},
  symbolic x coords={none,matched,mismatched}, xtick={none,matched,mismatched},
  xticklabels={{None},{Valid},{Out-of-date}},
  xticklabel style={font=\scriptsize}, xlabel={Experience}, xlabel style={font=\scriptsize, yshift=3pt},
  enlarge x limits=0.25, axis lines*=left, ymajorgrids, grid style={black!10},
  nodes near coords, point meta=explicit symbolic, nodes near coords style={font=\footnotesize\bfseries, yshift=2pt},
  error bars/y dir=both, error bars/y explicit, error bars/error bar style={line width=0.5pt},
  error bars/error mark options={rotate=90, mark size=1.5pt}]
\addplot[fill=black!35, draw=none, bar shift=0pt, nodes near coords style={text=black!75}] coordinates {(none,0.6319) +- (0,0.0074) [0.63]};
\addplot[fill=okgreen, draw=none, bar shift=0pt, nodes near coords style={text=black!75}] coordinates {(matched,0.9965) +- (0,0.0023) [1.00]};
\addplot[fill=badred, draw=none, bar shift=0pt, nodes near coords style={text=badred}] coordinates {(mismatched,0.2289) +- (0,0.0079) [0.23]};
\end{axis}
\end{tikzpicture}
\end{minipage}
\caption{Experience rot. (a) Deployed interfaces in which one field has opposite meanings, and in which changes need not
produce errors: Todoist's API reference \citep{todoist2026api} and two Todoist MCP servers
\citep[A:][B: \citealp{arora2025todoistmcp}]{greirson2025mcptodoist}; Python, JavaScript and POSIX conventions
\citep{python2026datetime,mdn2026getday,opengroup2018crontab}; the MCP specification \citep{mcp2025tools}; SilentProbe
\citep{li2026silentprobe}. (b) Two runs of DeepSeek-V4-Flash on the same final task with the same tool definition. Without
experience, the agent uses the field the definition lists. With experience from version A, it repeats its earlier
call; the service accepts the call, ignores the unknown field, and the agent reports success. (c) Success on the final task without experience, with valid experience, and with experience from before the
change, when the documentation gives the range of a value but not its meaning (incomplete documentation; six models, four changes;
Section~\ref{sec:effect}). Error bars are one standard error.}
\label{fig:example}
\end{figure*}

Agents are increasingly built to learn from experience. They keep conversation histories, write notes to memory files,
retrieve records of past tasks, and save scripts as reusable skills
\citep{packer2023memgpt,shinn2023reflexion,zhao2024expel,wang2024awm,wang2023voyager,chhikara2025mem0,zhou2026externalization}.
The idea is simple. An agent that has called a tool successfully once does not need to look up how to call it again.

This only works if the tool still works the way it did. Tools change. A field is renamed, a set of allowed values is
replaced, or the meaning of a number is reversed. Interfaces in use today already disagree on such details: in Todoist's
API reference and in two MCP servers for Todoist, task priority 1 is the highest in some places and 4 in others
(Figure~\ref{fig:example}a). Many changes are also silent. An MCP server can replace its tool definitions during a
session, and some services accept a request in the old form and ignore the fields they do not know
\citep{li2026silentprobe}. A survey of agent memory names such changes as an open problem \citep{du2026memory}, and
adaptation to them has not been tested \citep{qian2026temporal}.

We find that experienced agents fail where novices succeed. Figure~\ref{fig:example}b shows one case. An agent
schedules an event and puts the participants in \code{attendees}, as the service requires. The service then renames the
field to \code{guests}. With the next request, the agent receives a tool definition that lists \code{guests}. It writes
\code{attendees} anyway. The service accepts the call and stores an event without participants, and the agent tells
the user that the event is scheduled. An agent without the earlier task uses \code{guests} and succeeds. The pattern
holds across models and changes: after a silent change, experienced agents fail on most tasks, while agents new to the
task mostly succeed. Figure~\ref{fig:example}c shows this when the documentation is incomplete. Valid experience lifts
success from 63\% to 100\%. Out-of-date experience drops it to 23\%, far below agents without experience.

We call this failure \emph{experience rot}: an agent's own earlier tool use, correct when it was made, becomes wrong
after a silent change, and the agent reuses it anyway. The underlying behaviour is known. Agents reproduce the outputs
of similar past tasks, which \citet{xiong2025memory} call experience-following. We ask what this behaviour costs when
tools change, why agents do not notice, and what helps.

We study these questions in a controlled setting. Each service has two versions that differ in one interface detail.
An agent first completes a task on one version. The final task then runs on the same version (valid experience), on
the other version (out-of-date experience), or without the earlier task (no experience). The current documentation is
always one search away. With six models, we vary what the documentation says, where the experience is stored, how
much of it there is, what is added to the final task, and who wrote the earlier task. Most conditions were
preregistered. We report five findings.

\begin{enumerate}[leftmargin=*,itemsep=1pt,topsep=2pt]
\item \textbf{Out-of-date experience is worse than no experience} (Section~\ref{sec:effect}). When the documentation is
incomplete, experienced agents fall far below novices (Figure~\ref{fig:example}c). When it is complete, novices nearly
always succeed and experienced agents still fail on most tasks. Nearly every failed run ends with a report that the
task is done.
\item \textbf{Experienced agents stop checking} (Section~\ref{sec:checking}). They copy their earlier call instead of
reading the documentation, even when the tool definition in front of them shows the change. Lookups before the first
write drop to at most 6\% of runs, and agents talk about valid and out-of-date experience in the same way.
\item \textbf{Every form of memory is affected, and more experience makes it worse} (Section~\ref{sec:memory}). The
failure appears with conversation history, memory files, retrieved notes and skills, even when the task cannot be done
without the memory. It grows with the number of earlier tasks.
\item \textbf{Simple remedies fall short} (Section~\ref{sec:remedies}). Announcing the update barely helps. An
instruction to read the documentation helps only when it sits next to the request and the documentation contains the
answer; otherwise it does nothing or makes things worse. Checking the result after writing works, at more than twice
the model calls.
\item \textbf{Hand-written histories hide the problem} (Section~\ref{sec:authored}). Studies often give agents a
history written by the researchers. One check step in such a history removes most of the cost, because agents copy the
check along with the call.
\end{enumerate}

Our contributions are a controlled design that isolates the effect of an agent's own experience under silent tool
changes, the evidence above, and its consequences for memory systems, tool design and the evaluation of agent memory
(Section~\ref{sec:discussion}).

\section{Related Work}
\label{sec:related}

\paragraph{Following one's own experience.}
\citet{xiong2025memory} show that agents with a memory of past executions produce outputs similar to those of the
retrieved records when the task inputs are similar, and call this experience-following; errors in stored experience
then propagate to new tasks. In multi-turn settings, models imitate their own earlier responses
\citep[conversational inertia;][]{wan2026inertia}, keep to earlier reasoning when later turns bring corrections
\citep[contextual inertia;][]{chen2026contextual}, stay with early assumptions \citep{laban2025lost}, and make more
errors after seeing their own errors \citep{sinha2025illusion}. Older information in context also interferes with
updated values \citep{wang2025pi}. These works vary the experience or the conversation while the environment stays
fixed. We hold the agent's experience fixed in content and change the environment underneath it, and we measure whether
the agent checks the environment before it reuses the experience.

\paragraph{Changing tools.}
Most tool-use benchmarks keep the tools fixed \citep{yao2024taubench,lu2024toolsandbox,trivedi2024appworld}.
MCPEvol-Bench applies mutation operators to 123 real MCP servers and finds lower success on the evolved servers; adding
reflection, planning or memory modules reduces the drop \citep{liu2026mcpevol}. ProEvolve generates linked versions of
an environment for the same tasks; traces or summaries from earlier versions reduce the number of tool calls but do not
raise success, and lower it in several settings \citep{li2026world}. From an analysis of failures, its authors suggest
that agents with summaries of earlier versions hold outdated beliefs about which tools exist instead of consulting the
current tool list. We test this kind of explanation with controlled manipulations and measure the lookups directly.
ToolEVO trains models to adapt to API changes from explicit errors \citep{chen2025toolevo}. In SKILL.nb, workflow-memory
baselines lose 11--14 points when the web interface changes under the same tasks \citep{elhattami2026skillnb}. The
set-shifting test asks whether agents keep using a tool group that has stopped being reliable \citep{ye2026setshifting},
and ScrambleToolBench finds belief inertia when tool identifiers are permuted \citep{toh2026scramble}. MemDrift shows
that preferences stored in memory shift tool-call parameters in contexts where they do not apply
\citep{dabas2026tooldrift}. In code generation, models use deprecated APIs \citep{wang2025libevolution}, have difficulty
writing code for a specified library version \citep{islah2024gitchameleon}, and are hard to update with knowledge of
changed API functions \citep{liu2024codeupdatearena}.

\paragraph{Stale memory.}
STALE asks whether language models and memory systems recognise that a stored memory is no longer valid
\citep{chao2026stale}. \citet{fan2026statemem} describe state drift, where an agent acts on a stale version of a fact
that is present in its context. \citet{nakayashiki2026stale} give agents inherited memories with a superseded
constraint: under a budget of two verification actions out of six, agents rarely check the source record, while with a
budget of four they check it in 88.7\% of cases. \citet{chen2026compliance} study how agents consume conflicting memory,
and \citet{wang2026memtrap} study reasoning fixation and belief distortion caused by memories (both papers call their
benchmark MemTrapBench). \citet{hu2026continual} find that detailed trajectories in memory can hinder adaptation to
new tasks, while abstract procedural memories transfer more reliably. Most of these studies
write the memories themselves. We use the agent's own experience and show that hand-written histories change the
measured cost (Section~\ref{sec:authored}).

\paragraph{Information in view and silent failures.}
\citet{englander2026explore} find that agents rarely act on useful information they encounter in the environment, but
use it when it is placed in the task prompt (pass@1 rises from 55.88 to 76.25). We find a related pattern within the
context: documentation attached to the task message is used, documentation in the system prompt or the tool
description much less. \citet{cheng2025temporal} study when agents call a tool again instead of reusing earlier results.
Agents over-trust tool outputs that look valid \citep{sun2024toolsfail}. In production APIs, 44 of 61 constraints that
are stated only in prose fail silently, against none of 111 machine-checkable constraints, and four of five endpoints
ignore an undeclared parameter \citep{li2026silentprobe}. Implicit failures lower recovery by 37\% relative to explicit
errors \citep{zhu2026toolmaze}, and agents often report success on tasks they did not complete
\citep{advani2026falsesuccess}. Our silent changes follow these observations: unknown fields are ignored, and the
invisible changes are stated only in prose.

\paragraph{A human parallel.}
In the Einstellung effect, people who have solved problems with one method keep using it when a simpler method is
available \citep{luchins1942}, and chess experts who know a familiar solution keep attending to it while they report
looking for a better one \citep{bilalic2008}. Studies of language models have used this framing for reasoning
benchmarks \citep{haq2025mentalset,kim2025inflexible}. We note the resemblance in behaviour only and make no claim
about a shared mechanism.

\section{Setup}
\label{sec:setup}

The design compares agents that differ in one respect: whether their experience comes from the current version of a
tool or from an earlier one.

\subsection{Services and changes}
\label{sec:services}

We built three small services: a calendar, a task list, and a mail client. Each service has two versions, A and B,
that differ in one interface detail (Table~\ref{tab:changes}). The agent reaches a service through tools with typed
definitions, in the form used by function calling and by the Model Context Protocol: each tool has a name, a
description, and a JSON schema for its arguments. Each service also offers read tools that return the stored state
(for example \code{calendar\_get\_event}) and a \code{docs\_search} tool that returns the reference documentation of
the running version (Appendix~\ref{app:services}).

\begin{table}[t]
\centering
\scriptsize
\setlength{\tabcolsep}{3pt}
\resizebox{\linewidth}{!}{%
\begin{tabular}{llll}
\toprule
Change & Version A & Version B & In definition \\
\midrule
Field name & \code{attendees} & \code{guests} & yes \\
Structure & \code{end\_time} & \code{duration\_minutes} & yes \\
Value & \code{urgent}, \dots, \code{low} & \code{p0}, \dots, \code{p3} & yes \\
\midrule
Priority direction & 1 is highest & 4 is highest & no \\
Weekday numbering & 0 = Monday & 0 = Sunday & no \\
\midrule
Month numbering$^*$ & 1 = January & 0 = January & no \\
Duration unit$^*$ & minutes & seconds & no \\
\bottomrule
\end{tabular}}
\caption{The changes. In the first three, a field or value changes, and a typed tool definition lists the fields or
values of the running version. In the others, a number keeps its range but changes its meaning, and the tool
definition is the same in both versions (for example an integer from 1 to 4). $^*$Used only with incomplete
documentation.}
\label{tab:changes}
\end{table}

The tool definitions always follow the running version: an earlier task on version A shows the definitions of A, and
a final task on version B shows those of B. Changes are silent. A request that is valid only in the other version is
accepted with the same response as a correct request, \code{\{"ok": true, "result": \{"accepted": true, \dots\}\}}.
An unknown field is ignored, an unknown value leaves the stored state unchanged, and an integer takes the meaning of the
running version. In Figure~\ref{fig:example}b, the event is stored without participants, and the response does not say
so. The read tools show the stored state in plain terms (``Highest'', ``Tuesday'', the end time of an event), so an
agent can always check what its write did.

\paragraph{What the documentation says.}
With \emph{complete documentation}, the documentation of each version gives the fields, the values and what the values
mean. The agent can then learn of a change in one of three ways. For a changed field or value, the write tool either
takes an untyped \code{payload}, so that only the documentation gives the fields, or has a typed definition that
lists the fields or values of the running version. For a changed meaning, the definition is the same in both versions,
and only the documentation says what the number means. With \emph{incomplete documentation}, the documentation gives
only the range of a value, for example ``\code{priority} is an integer from 1 to 4'', and not what it means. Real
documentation is often like this (Figure~\ref{fig:example}a).

\subsection{Episodes}

An episode has an earlier task and a final task of the same kind, for example scheduling two different events. The
agent does the earlier task live, with the tools and tool definitions of the version it runs on. An attempt counts only
if the stored state matches the request; failed attempts are discarded and repeated, up to ten times. The final task
then follows as the next message in the same conversation. There are three arms. With \emph{valid experience}, the
earlier task ran on the same version as the final task. With \emph{out-of-date experience}, it ran on the other
version. With \emph{no experience}, there is no earlier task. Each change is run in both directions, A to B and B to A,
and the same task instances are used in all arms. The valid and out-of-date arms therefore differ only in the version
on which the agent gained its experience. With incomplete documentation, the earlier request says what the values mean
on its version, for example ``(For this service, priority 4 is the highest and 1 is the lowest.)''; the final request
does not.

Later sections change one part of this design at a time: where the experience is stored and how much of it there is
(Section~\ref{sec:memory}), what is added to the final task (Section~\ref{sec:remedies}), and who wrote the earlier task
(Section~\ref{sec:authored}). Appendix~\ref{app:conditions} lists all conditions.

\paragraph{Measures.}
A program checks the stored state after the final task. \emph{Success} means that the state matches the request. The
\emph{cost} of out-of-date experience is the success rate with valid experience minus the success rate with
out-of-date experience. We also record whether the agent searched the documentation before its first write
(\emph{lookup}), whether its first write uses the usage of the other version (\emph{old usage}; for example
\code{attendees} on version B, or 1 for the highest priority when 4 is the highest), whether it called a read tool
after its first write (\emph{read-back}), the number of model calls, and whether the final reply reports the task as
done (Appendix~\ref{app:reports}).

\subsection{Models and statistics}

We test six models: GLM-5.3-Flash, Muse-Glimmer-30B, DeepSeek-V4-Flash, MiniMax-M3,
Qwen3.8-Flash, and Inkling (Appendix~\ref{app:models}). Each model runs the same simple agent loop with temperature 0.7
and at most 16 model calls per task. Of 12,624 episodes, 136 are excluded: 9 because a model request failed and 127 because the
earlier task never succeeded, most of them on the weekday change (Appendix~\ref{app:calibration}).

Estimates are pooled over models, changes and directions. We report each estimate $\pm$ one standard error from a
cluster bootstrap over task instances: instances are resampled within each change and direction, jointly for all models
and conditions, with 10,000 replicates. We call a difference positive or negative when its 95\% percentile interval
excludes zero. For most conditions with complete documentation, the design, the predictions, the exclusion rules and
the analysis code were preregistered before the run, and all preregistered predictions held (Appendix~\ref{app:prereg}).
The conditions with incomplete documentation were not preregistered.

\section{Experienced Agents Fail Where Novices Succeed}
\label{sec:effect}

\begin{figure*}[t]
\centering
\armlegend\\[1mm]
\begin{minipage}[t]{0.60\textwidth}\centering
% Generated by prior_drift/paper_analysis/paper_figs.py. Do not edit by hand.
\begin{tikzpicture}
\begin{axis}[paperbars, width=\linewidth, height=4.3cm, bar width=7pt, ymax=1.18, ylabel={Success on the final task}, xticklabel style={text width=1.9cm},
  xmin=-0.5, xmax=3.5, xtick={0,1,2,3},
  xticklabels={{Meaning of\\a number},{Field, untyped\\tool},{Field, in the\\definition},{Meaning of\\a number}}]
\addplot[armnone] coordinates {(0,0.6319) +- (0,0.0074) (1,1.0000) +- (0,0.0000) (2,0.9953) +- (0,0.0042) (3,0.8750) +- (0,0.0117)};
\addplot[armvalid] coordinates {(0,0.9965) +- (0,0.0023) (1,1.0000) +- (0,0.0000) (2,1.0000) +- (0,0.0000) (3,1.0000) +- (0,0.0000)};
\addplot[armood, nodes near coords, point meta=explicit symbolic, nodes near coords style={font=\tiny, text=badred, yshift=1pt}] coordinates {(0,0.2289) +- (0,0.0079) [0.23] (1,0.2605) +- (0,0.0125) [0.26] (2,0.4861) +- (0,0.0178) [0.49] (3,0.4179) +- (0,0.0197) [0.42]};
\draw[black!30, dashed] (axis cs:0.5,0) -- (axis cs:0.5,1.15);
\node[font=\scriptsize\itshape, text=black!65, anchor=south] at (axis cs:0,1.13) {Incomplete docs};
\node[font=\scriptsize\itshape, text=black!65, anchor=south] at (axis cs:2,1.13) {Complete docs};
\end{axis}
\end{tikzpicture}\\[-0.5mm]
{\footnotesize (a) Success on the final task}
\end{minipage}\hfill
\begin{minipage}[t]{0.38\textwidth}\centering
% Generated by prior_drift/paper_analysis/paper_figs.py. Do not edit by hand.
\begin{tikzpicture}
\begin{axis}[xbar, clip=false, width=0.6\linewidth, height=4.3cm, xmin=0, xmax=1.12,
  bar width=8pt, y dir=reverse, ymin=-0.6, ymax=3.6, axis lines*=left, xmajorgrids, grid style={black!10},
  tick label style={font=\scriptsize}, label style={font=\scriptsize}, xtick={0,0.5,1}, xticklabels={0,50\%,100\%},
  ytick={0,1,2,3}, yticklabels={{Incomplete docs,\\meaning of a number},{Complete docs,\\field, untyped tool},{Complete docs,\\field, in the definition},{Complete docs,\\meaning of a number}}, ytick style={draw=none},
  yticklabel style={align=right, font=\scriptsize}, xlabel={Failed runs reported as done},
  nodes near coords, point meta=explicit symbolic, nodes near coords style={font=\tiny, text=black!75}]
\addplot[fill=badred!55, draw=none] coordinates {(0.9247,0) [92\%] (1.0000,1) [100\%] (1.0000,2) [100\%] (0.9872,3) [99\%]};
\end{axis}
\end{tikzpicture}\\[-0.5mm]
{\footnotesize (b) Failed runs, out-of-date experience}
\end{minipage}
\caption{(a) Success on the same final tasks with no, valid and out-of-date experience, $\pm$ one standard error.
``Field'': a renamed field, a changed structure or a changed set of values, with an untyped write tool or with a typed
definition that lists the current fields. ``Meaning of a number'': a number keeps its range and changes its meaning.
(b) Share of failed runs with out-of-date experience whose final reply tells the user that the task is done. Six
models; numbers in Table~\ref{tab:main}.}
\label{fig:effect}
\end{figure*}
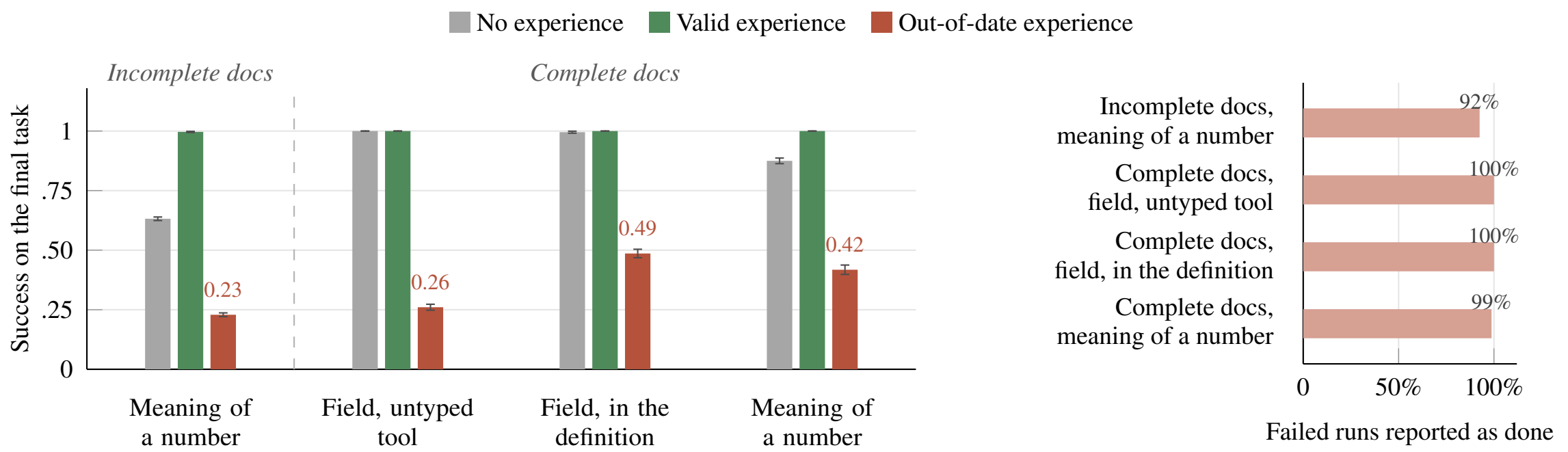

Figure~\ref{fig:effect} compares the three arms on the same final tasks.

\paragraph{With incomplete documentation, experience is what makes agents succeed.}
Without experience, agents succeed in 63\% of tasks. They follow their own default and succeed when the running
version happens to match it, for example in every run where January is month 1 and in 40\% where it is month 0
(Appendix~\ref{app:undocumented}). With valid experience, success is 100\%.

\paragraph{After a change, the same experience is worse than none.}
With out-of-date experience, success falls to 23\%, $0.40 \pm 0.01$ below agents without experience. Every first
write uses the meaning from the earlier version. The drop appears for each of the four changes, with success between
15\% and 31\% (Table~\ref{tab:undoc-change}).

\paragraph{Experienced agents fail even when the answer is in front of them.}
With complete documentation, agents without experience succeed in 88--100\% of tasks, and agents with valid experience
in all of them. With out-of-date experience, success falls to 26--49\%. It is 49\% when the tool definition in front of
the agent lists the new field: the definition says \code{guests}, and the agent writes \code{attendees}
(Figure~\ref{fig:example}b).

\paragraph{Agents report the failures as successes.}
The service accepts the outdated call, and the agent tells the user that the task is done (Figure~\ref{fig:effect}b).
This happens in 347 of 348 failed runs with complete documentation, for example ``The event `Security Review' has been
scheduled \dots{} with Paul and Maya as attendees'' (MiniMax-M3) for an event stored without participants. With
incomplete documentation the share is 92\% (Appendix~\ref{app:reports}).

\section{Experienced Agents Stop Checking}
\label{sec:checking}

\begin{figure*}[t]
\centering
\armlegend\\[1mm]
\begin{minipage}[t]{0.47\textwidth}\centering
% Generated by prior_drift/paper_analysis/paper_figs.py. Do not edit by hand.
\begin{tikzpicture}
\begin{axis}[paperbars, width=\linewidth, height=4.3cm, bar width=6pt, ymax=1.18, ylabel={Lookup before the first write}, xticklabel style={text width=1.45cm},
  xmin=-0.5, xmax=3.5, xtick={0,1,2,3},
  xticklabels={{Meaning of\\a number},{Field, untyped\\tool},{Field, in the\\definition},{Meaning of\\a number}}]
\addplot[armnone] coordinates {(0,0.3507) +- (0,0.0000) (1,1.0000) +- (0,0.0000) (2,0.2512) +- (0,0.0000) (3,0.7014) +- (0,0.0000)};
\addplot[armvalid] coordinates {(0,0.0176) +- (0,0.0000) (1,0.0139) +- (0,0.0000) (2,0.0000) +- (0,0.0000) (3,0.0149) +- (0,0.0000)};
\addplot[armood] coordinates {(0,0.0141) +- (0,0.0000) (1,0.0093) +- (0,0.0000) (2,0.0648) +- (0,0.0000) (3,0.0000) +- (0,0.0000)};
\draw[black!30, dashed] (axis cs:0.5,0) -- (axis cs:0.5,1.15);
\node[font=\scriptsize\itshape, text=black!65, anchor=south] at (axis cs:0,1.13) {Incomplete docs};
\node[font=\scriptsize\itshape, text=black!65, anchor=south] at (axis cs:2,1.13) {Complete docs};
\end{axis}
\end{tikzpicture}\\[-0.5mm]
{\footnotesize (a) Lookup before the first write}
\end{minipage}\hfill
\begin{minipage}[t]{0.25\textwidth}\centering
% Generated by prior_drift/paper_analysis/paper_figs.py. Do not edit by hand.
\begin{tikzpicture}
\begin{axis}[paperbars, width=\linewidth, height=4.3cm, bar width=8pt, ymax=0.8, ytick={0,0.2,0.4,0.6,0.8}, yticklabels={0,.2,.4,.6,.8}, ylabel={Share of runs}, xticklabel style={text width=1.3cm},
  xmin=-0.38, xmax=1.38, xtick={0,1},
  xticklabels={{Plans to\\check the\\docs},{Refers to\\the earlier\\task}}]
\addplot[armnone] coordinates {(0,0.6886) +- (0,0.0000) (1,0.0155) +- (0,0.0000)};
\addplot[armvalid] coordinates {(0,0.0590) +- (0,0.0000) (1,0.2359) +- (0,0.0000)};
\addplot[armood] coordinates {(0,0.0909) +- (0,0.0000) (1,0.2531) +- (0,0.0000)};
\end{axis}
\end{tikzpicture}\\[-0.5mm]
{\footnotesize (b) What the agent says}
\end{minipage}\hfill
\begin{minipage}[t]{0.25\textwidth}\centering
% Generated by prior_drift/paper_analysis/paper_figs.py. Do not edit by hand.
\begin{tikzpicture}
\begin{axis}[width=0.95\linewidth, height=4.3cm, xmin=-0.05, xmax=1.08, ymin=-0.05, ymax=1.08,
  clip=false, axis lines*=left, tick label style={font=\scriptsize}, label style={font=\scriptsize},
  xtick={0,0.5,1}, ytick={0,0.5,1}, xlabel={Read-back after the first write}, ylabel={Success, out-of-date},
  ymajorgrids, xmajorgrids, grid style={black!10}]
\draw[black!25] (axis cs:0,0) -- (axis cs:1,1);
\addplot[only marks, mark=*, mark size=1.6pt, slate] coordinates {(0.533,0.700) (0.233,0.250) (0.000,0.351) (0.000,0.052) (1.000,1.000) (0.000,0.382)};
\addplot[only marks, mark=o, mark size=1.8pt, badred, thick] coordinates {(0.552,0.396) (0.198,0.021) (0.000,0.000) (0.010,0.010) (1.000,0.927) (0.000,0.000)};
\node[font=\tiny, text=slate, anchor=west] at (axis cs:0.50,0.13) {$\bullet$ complete docs};
\node[font=\tiny, text=badred, anchor=west] at (axis cs:0.50,0.01) {$\circ$ incomplete docs};
\node[font=\tiny, text=black!70, anchor=east] at (axis cs:0.970,1.000) {Qwen};
\node[font=\tiny, text=black!70, anchor=west] at (axis cs:0.563,0.700) {GLM};
\node[font=\tiny, text=black!70, anchor=west] at (axis cs:0.263,0.250) {Muse};
\node[font=\tiny, text=black!70, anchor=west] at (axis cs:0.03,0.366) {\shortstack[l]{DeepSeek,\\Inkling}};
\node[font=\tiny, text=black!70, anchor=west] at (axis cs:0.03,0.112) {MiniMax};
\end{axis}
\end{tikzpicture}\\[-0.5mm]
{\footnotesize (c) By model}
\end{minipage}
\caption{(a) Share of runs that search the documentation before the first write. (b) Runs whose reasoning or reply
text before the first write plans to check the documentation or refers to the earlier task (complete documentation).
(c) Per model, read-back after the first write against success with out-of-date experience; filled: complete
documentation with typed definitions, open: incomplete documentation. The line marks equal values.}
\label{fig:checking}
\end{figure*}
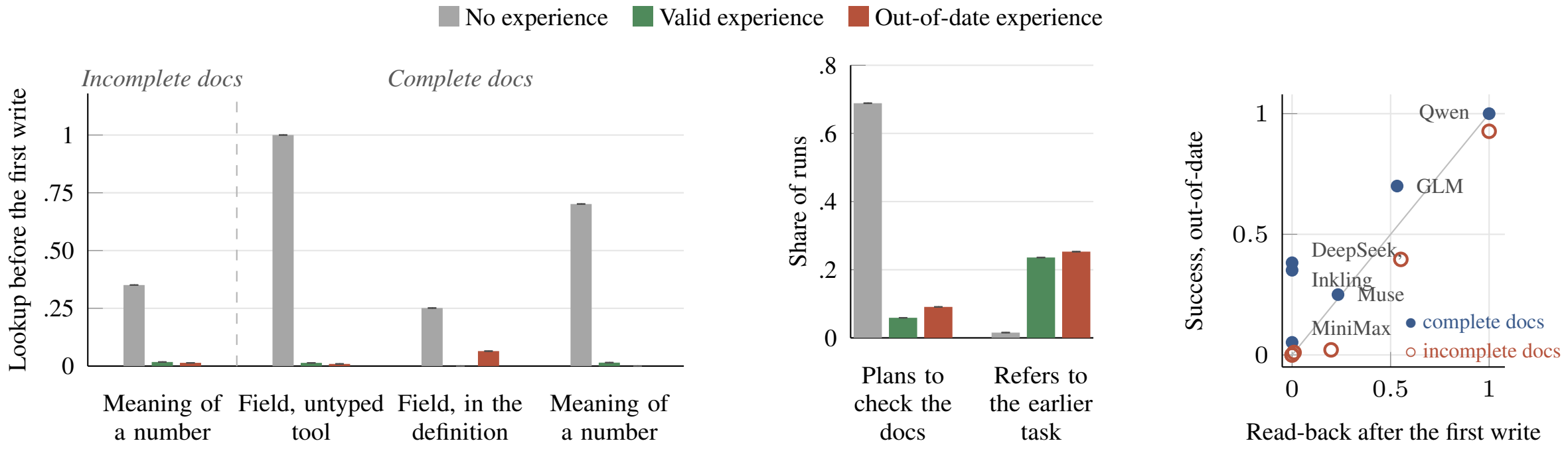

The current interface is one search away, and sometimes already in the tool definition. Experienced agents still fail
because they copy their earlier call without checking it.

\paragraph{They stop looking things up.}
Agents without experience search the documentation when they need it: in every run when only the documentation gives
the fields, and in 70\% of runs when a number changes its meaning (Figure~\ref{fig:checking}a). With experience, valid
or out of date, at most 6\% do. The earlier call alone is enough for this. In 135 of 144 earlier tasks with a renamed
field or a changed structure, the agent made no lookup, so no documentation of the old version was in the
conversation; the first write still repeated the old field in 61\% of these runs. Agents also copy details that the
task does not fix, such as the format of the start time (Appendix~\ref{app:copy}).

\paragraph{They do not tell valid from out-of-date experience.}
Before the first write, agents without experience often say that they will check the documentation. Experienced agents
instead refer to the earlier task, and they do so as often when their experience is out of date as when it is valid
(Figure~\ref{fig:checking}b). Typical statements are ``I already know from the previous request that the highest
priority is `urgent'{}'' (MiniMax-M3) and ``I could search docs again, but I already know the format from the
previous search'' (GLM-5.3-Flash). One GLM run notices that the current schema no longer lists
\code{duration\_minutes} and keeps it anyway: ``The tool accepted duration\_minutes before. I'll do the same.''

\paragraph{Agents that read back their result recover.}
Most failures follow one path: no lookup, old usage, no read-back. Runs that write the old usage and then read back the
stored state almost always see the error and correct it (94 of 97; Appendix~\ref{app:paths}). Across models, success
with out-of-date experience follows the read-back rate in both documentation settings (Figure~\ref{fig:checking}c).
Qwen3.8-Flash reads back after every write and nearly always recovers; with complete documentation, MiniMax-M3,
DeepSeek-V4-Flash and Inkling never read back.

\section{Every Form of Memory Is Affected, and More Experience Makes It Worse}
\label{sec:memory}

\begin{figure*}[t]
\centering
\armlegend\\[1mm]
\begin{minipage}[t]{0.58\textwidth}\centering
% Generated by prior_drift/paper_analysis/paper_figs.py. Do not edit by hand.
\begin{tikzpicture}
\begin{axis}[paperbars, width=\linewidth, height=4.3cm, bar width=7pt, ylabel={Success on the final task}, xticklabel style={text width=1.4cm},
  xmin=-0.5, xmax=3.5, xtick={0,1,2,3},
  xticklabels={{Conver-\\sation},{Memory\\file},{Retrieved\\memory},{Skill}}]
\addplot[armnone] coordinates {(0,0.0667) +- (0,0.0304) (1,0.0667) +- (0,0.0312) (2,0.1000) +- (0,0.0429) (3,0.0333) +- (0,0.0183)};
\addplot[armvalid] coordinates {(0,1.0000) +- (0,0.0000) (1,0.9887) +- (0,0.0064) (2,0.9607) +- (0,0.0112) (3,0.9111) +- (0,0.0144)};
\addplot[armood, nodes near coords, point meta=explicit symbolic, nodes near coords style={font=\tiny, text=badred, yshift=1pt}] coordinates {(0,0.4689) +- (0,0.0157) [0.47] (1,0.7175) +- (0,0.0254) [0.72] (2,0.7627) +- (0,0.0138) [0.76] (3,0.6111) +- (0,0.0234) [0.61]};
\end{axis}
\end{tikzpicture}\\[-0.5mm]
{\footnotesize (a) Where the experience is stored}
\end{minipage}\hfill
\begin{minipage}[t]{0.38\textwidth}\centering
% Generated by prior_drift/paper_analysis/paper_figs.py. Do not edit by hand.
\begin{tikzpicture}
\begin{axis}[paperbars, width=\linewidth, height=4.3cm, bar width=8pt, xlabel={Earlier tasks in the conversation},
  xmin=-0.5, xmax=2.5, xtick={0,1,2},
  xticklabels={{1},{3},{10}}]
\addplot[armvalid] coordinates {(0,1.0000) +- (0,0.0000) (1,0.9887) +- (0,0.0065) (2,0.9716) +- (0,0.0090)};
\addplot[armood, nodes near coords, point meta=explicit symbolic, nodes near coords style={font=\tiny, text=badred, yshift=1pt}] coordinates {(0,0.4600) +- (0,0.0135) [0.46] (1,0.3333) +- (0,0.0190) [0.33] (2,0.2629) +- (0,0.0189) [0.26]};
\end{axis}
\end{tikzpicture}\\[-0.5mm]
{\footnotesize (b) How much experience}
\end{minipage}
\caption{(a) Success on final tasks that need the earlier task, by where the experience is stored. (b) Success with
one, three or ten earlier tasks in the conversation. Five changes, six models, $\pm$ one standard error; numbers in
Tables~\ref{tab:memory} and~\ref{tab:authored-dose}.}
\label{fig:memory}
\end{figure*}
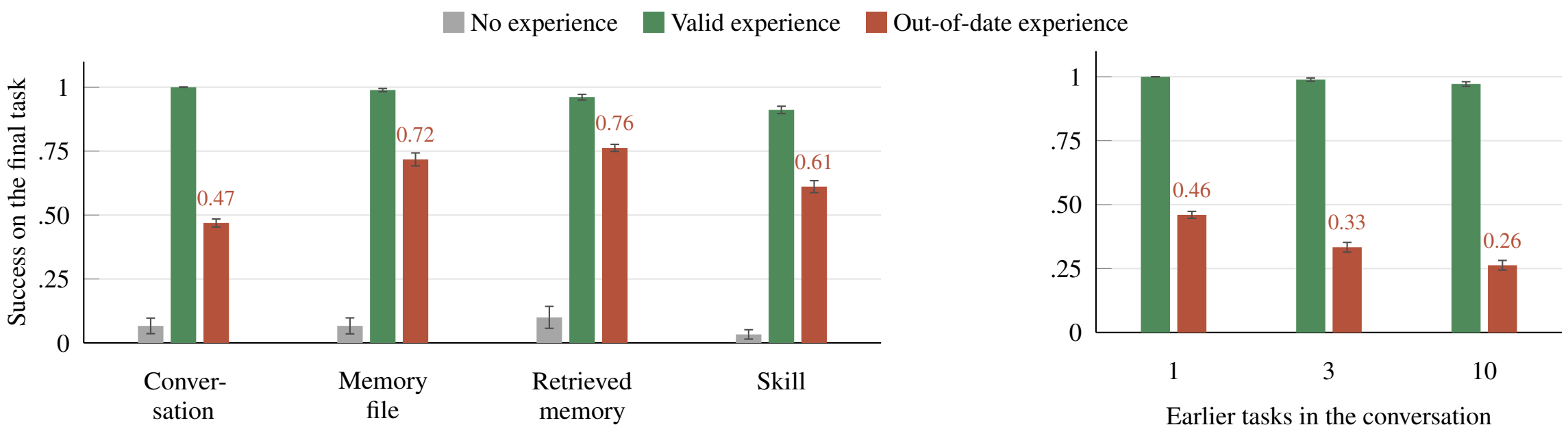

So far the earlier task stayed in the conversation, and the agent could do the final task without it. Agents in
practice keep experience in other forms, and they keep it because they need it. We therefore make the experience
necessary. Each earlier request contains information that the final request refers to, such as ``These two people are
my review panel'' and later ``with my review panel as participants''. Without the earlier task, agents succeed in
3--10\% of these final tasks. We store the experience in four ways: in the \emph{conversation}; in a \emph{memory
file} that the agent writes after the earlier task and that appears in the system prompt of a new session; as
\emph{retrieved memory}, the three entries of that note that best match the final request; and as a \emph{skill}, a
reusable Python script that the agent wrote and ran, together with the note.

\paragraph{The cost appears in every design.}
In every design, valid experience makes the final task possible, and out-of-date experience loses much of that gain
(Figure~\ref{fig:memory}a). The cost is 0.53 in the conversation and 0.20--0.30 in the other designs. The content of
the experience is carried over: with out-of-date experience, the first write has the right content, such as the right
participants, in 95--100\% of runs. What fails is the interface. The agents write the old usage into their notes:
60--73\% of the memory files record the field or value of the earlier version, and 92\% of the notes that come with a
skill. The note-based designs lose less than the conversation, and agents with a note search the documentation more
often (24--34\% against 2\%). We did not test why; \citet{hu2026continual} likewise find that abstract memories
transfer more reliably than detailed trajectories.

\paragraph{More earlier tasks, higher cost.}
With one, three or ten earlier tasks of the same kind in the conversation, success with out-of-date experience falls
from 46\% to 33\% and 26\%, while success with valid experience stays at 97--100\% (Figure~\ref{fig:memory}b). The
lookup rate falls from 4\% to 0\%. An agent that has used a tool more often relies on its experience more.

\section{Simple Remedies Fall Short}
\label{sec:remedies}

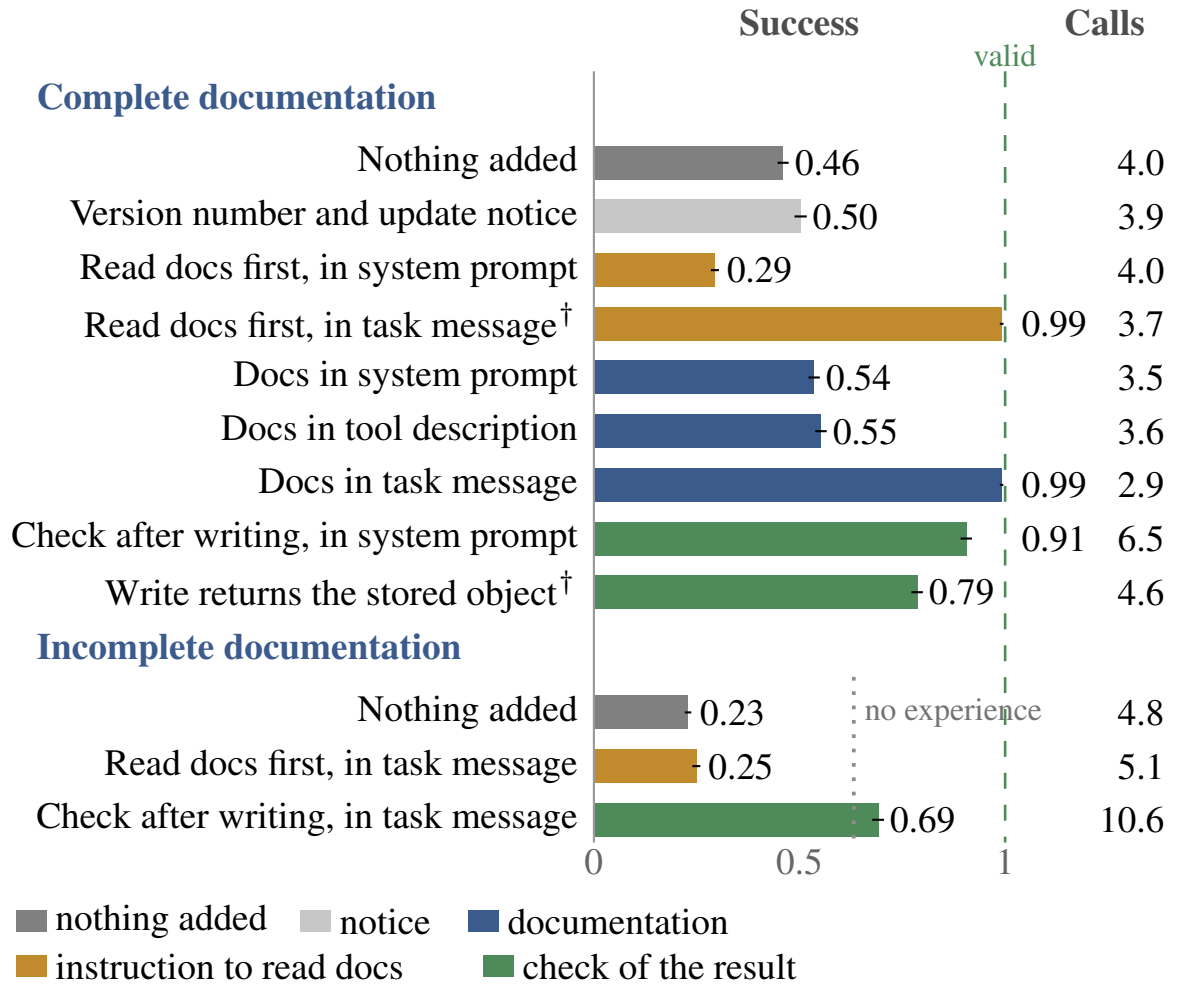
\begin{figure}[t]
\centering
% Generated by prior_drift/paper_analysis/fig_noticing.py. Do not edit by hand.
\begin{tikzpicture}[font=\scriptsize, x=1cm, y=1cm]
\node[anchor=south, font=\scriptsize\bfseries, text=black!70] at (5.180,0.050) {Success};
\node[anchor=south east, font=\scriptsize\bfseries, text=black!70] at (7.600,0.050) {Calls};
\node[anchor=west, font=\scriptsize\bfseries, text=slate] at (0,-0.240) {Complete documentation};
\node[anchor=east, inner sep=0pt] at (3.720,-0.655) {Nothing added};
\fill[black!50] (3.820,-0.765) rectangle (5.071,-0.545);
\draw[line width=0.4pt] (5.035,-0.655) -- (5.108,-0.655);
\node[anchor=west, inner sep=1.5pt] at (5.091,-0.655) {0.46};
\node[anchor=east, inner sep=0pt] at (7.600,-0.655) {4.0};
\node[anchor=east, inner sep=0pt] at (3.720,-1.010) {Version number and update notice};
\fill[black!22] (3.820,-1.120) rectangle (5.188,-0.900);
\draw[line width=0.4pt] (5.147,-1.010) -- (5.228,-1.010);
\node[anchor=west, inner sep=1.5pt] at (5.208,-1.010) {0.50};
\node[anchor=east, inner sep=0pt] at (7.600,-1.010) {3.9};
\node[anchor=east, inner sep=0pt] at (3.720,-1.365) {Read docs first, in system prompt};
\fill[amber] (3.820,-1.475) rectangle (4.621,-1.255);
\draw[line width=0.4pt] (4.594,-1.365) -- (4.647,-1.365);
\node[anchor=west, inner sep=1.5pt] at (4.641,-1.365) {0.29};
\node[anchor=east, inner sep=0pt] at (7.600,-1.365) {4.0};
\node[anchor=east, inner sep=0pt] at (3.720,-1.720) {Read docs first, in task message$^\dagger$};
\fill[amber] (3.820,-1.830) rectangle (6.517,-1.610);
\draw[line width=0.4pt] (6.505,-1.720) -- (6.529,-1.720);
\node[anchor=west, inner sep=1.5pt] at (6.590,-1.720) {0.99};
\node[anchor=east, inner sep=0pt] at (7.600,-1.720) {3.7};
\node[anchor=east, inner sep=0pt] at (3.720,-2.075) {Docs in system prompt};
\fill[slate] (3.820,-2.185) rectangle (5.276,-1.965);
\draw[line width=0.4pt] (5.239,-2.075) -- (5.314,-2.075);
\node[anchor=west, inner sep=1.5pt] at (5.296,-2.075) {0.54};
\node[anchor=east, inner sep=0pt] at (7.600,-2.075) {3.5};
\node[anchor=east, inner sep=0pt] at (3.720,-2.430) {Docs in tool description};
\fill[slate] (3.820,-2.540) rectangle (5.322,-2.320);
\draw[line width=0.4pt] (5.284,-2.430) -- (5.360,-2.430);
\node[anchor=west, inner sep=1.5pt] at (5.342,-2.430) {0.55};
\node[anchor=east, inner sep=0pt] at (7.600,-2.430) {3.6};
\node[anchor=east, inner sep=0pt] at (3.720,-2.785) {Docs in task message};
\fill[slate] (3.820,-2.895) rectangle (6.517,-2.675);
\draw[line width=0.4pt] (6.505,-2.785) -- (6.528,-2.785);
\node[anchor=west, inner sep=1.5pt] at (6.590,-2.785) {0.99};
\node[anchor=east, inner sep=0pt] at (7.600,-2.785) {2.9};
\node[anchor=east, inner sep=0pt] at (3.720,-3.140) {Check after writing, in system prompt};
\fill[okgreen] (3.820,-3.250) rectangle (6.282,-3.030);
\draw[line width=0.4pt] (6.245,-3.140) -- (6.320,-3.140);
\node[anchor=west, inner sep=1.5pt] at (6.590,-3.140) {0.91};
\node[anchor=east, inner sep=0pt] at (7.600,-3.140) {6.5};
\node[anchor=east, inner sep=0pt] at (3.720,-3.495) {Write returns the stored object$^\dagger$};
\fill[okgreen] (3.820,-3.605) rectangle (5.958,-3.385);
\draw[line width=0.4pt] (5.920,-3.495) -- (5.996,-3.495);
\node[anchor=west, inner sep=1.5pt] at (5.978,-3.495) {0.79};
\node[anchor=east, inner sep=0pt] at (7.600,-3.495) {4.6};
\node[anchor=west, font=\scriptsize\bfseries, text=slate] at (0,-3.875) {Incomplete documentation};
\node[anchor=east, inner sep=0pt] at (3.720,-4.290) {Nothing added};
\fill[black!50] (3.820,-4.400) rectangle (4.443,-4.180);
\draw[line width=0.4pt] (4.421,-4.290) -- (4.464,-4.290);
\node[anchor=west, inner sep=1.5pt] at (4.463,-4.290) {0.23};
\node[anchor=east, inner sep=0pt] at (7.600,-4.290) {4.8};
\node[anchor=east, inner sep=0pt] at (3.720,-4.645) {Read docs first, in task message};
\fill[amber] (3.820,-4.755) rectangle (4.498,-4.535);
\draw[line width=0.4pt] (4.471,-4.645) -- (4.524,-4.645);
\node[anchor=west, inner sep=1.5pt] at (4.518,-4.645) {0.25};
\node[anchor=east, inner sep=0pt] at (7.600,-4.645) {5.1};
\node[anchor=east, inner sep=0pt] at (3.720,-5.000) {Check after writing, in task message};
\fill[okgreen] (3.820,-5.110) rectangle (5.700,-4.890);
\draw[line width=0.4pt] (5.661,-5.000) -- (5.739,-5.000);
\node[anchor=west, inner sep=1.5pt] at (5.720,-5.000) {0.69};
\node[anchor=east, inner sep=0pt] at (7.600,-5.000) {10.6};
\draw[okgreen, dashed, line width=0.5pt] (6.540,-0.050) -- (6.540,-5.150);
\draw[black!45, dotted, line width=0.7pt] (5.539,-4.055) -- (5.539,-5.180);
\draw[black!40, line width=0.4pt] (3.820,-0.050) -- (3.820,-5.150);
\node[anchor=north, text=black!60, inner sep=1pt] at (3.820,-5.150) {0};
\node[anchor=north, text=black!60, inner sep=1pt] at (5.180,-5.150) {0.5};
\node[anchor=north, text=black!60, inner sep=1pt] at (6.540,-5.150) {1};
\node[anchor=south, text=okgreen, inner sep=1pt, font=\tiny] at (6.540,-0.060) {valid};
\fill[black!50] (0.000,-5.740) rectangle (0.200,-5.600);
\node[anchor=west, inner sep=1.5pt] at (0.200,-5.670) {nothing added};
\fill[black!22] (1.880,-5.740) rectangle (2.080,-5.600);
\node[anchor=west, inner sep=1.5pt] at (2.080,-5.670) {notice};
\fill[slate] (2.990,-5.740) rectangle (3.190,-5.600);
\node[anchor=west, inner sep=1.5pt] at (3.190,-5.670) {documentation};
\fill[amber] (0.000,-6.040) rectangle (0.200,-5.900);
\node[anchor=west, inner sep=1.5pt] at (0.200,-5.970) {instruction to read docs};
\fill[okgreen] (3.090,-6.040) rectangle (3.290,-5.900);
\node[anchor=west, inner sep=1.5pt] at (3.290,-5.970) {check of the result};
\node[anchor=west, text=black!55, inner sep=1pt, font=\tiny] at (5.569,-4.290) {no experience};
\end{tikzpicture}
\caption{Success with out-of-date experience and model calls per task, by what is added to the final task. Upper
block: complete documentation (five changes; $^\dagger$run separately on the same task instances, where nothing added
gives 0.45). Lower block: incomplete documentation (four changes). Six models; lines mark $\pm$ one standard error.
The dashed line marks success with valid experience (2.9 and 3.1 calls), the dotted line success without experience
with incomplete documentation.}
\label{fig:noticing}
\end{figure}

If agents copy their experience only because they have no reason to doubt it, a simple addition to the context should
be enough. We tested three kinds of additions to the final task: telling the agent that the service has changed,
pointing it to the documentation, and making it check the result (Figure~\ref{fig:noticing};
Appendix~\ref{app:context}). In each condition the agent has out-of-date experience in the conversation. With nothing
added, success is 46\% with complete documentation and 23\% with incomplete documentation.

\paragraph{Telling the agent that the service changed barely helps.}
A version number in every tool description and an update notice in the final task (``The service was updated on
2026-03-01; see the reference documentation for details.'') raise success only to 50\%. Recording the service version
in a memory file lowers the cost only from 0.27 to 0.20 (Table~\ref{tab:memory}).

\paragraph{Pointing to the documentation works only next to the request, and only if the documentation has the answer.}
An instruction in the system prompt to check the documentation before writing lowers success to 29\%: the agents follow
it in the earlier task (98\%) but not in the final task (4\%). Appended to the final task message, the same instruction
makes every agent search, and success rises to 99\%. The documentation itself behaves the same way: in the task message
it gives 99\%, in the system prompt or the tool description 54--55\%. Placed next to a memory file, it lowers the cost
from 0.27 to 0.10. With incomplete documentation, the instruction in the task message again makes every agent search,
but the documentation does not contain the answer, and success stays at 25\%. Even when the experience is still valid,
the instruction raises the model calls per task from 2.9 to 3.6 and the input tokens by 40\%
(Appendix~\ref{app:undocumented}).

\paragraph{Checking the result works, at more than twice the model calls.}
An instruction to read back the stored state before reporting completion, and to correct any mismatch, raises success
to 91\% with complete documentation and to 69\% with incomplete documentation. The agents still write the old usage
first, then read back and correct it, at 6.5 and 10.6 model calls per task against 2.9 and 3.1 with valid experience.
A service can also show the result without being asked. When every write returns the stored object, success rises
from 45\% to 79\% at 4.6 calls per task. In 59 of the 77 remaining failures, the agent receives a stored object that
shows the wrong state and does not write again.

\section{Hand-Written Histories Hide the Problem}
\label{sec:authored}

\begin{figure}[t]
\centering
{\scriptsize\armkey{okgreen}~Valid experience\quad\armkey{badred}~Out-of-date experience}\\[1mm]
% Generated by prior_drift/paper_analysis/paper_figs.py. Do not edit by hand.
\begin{tikzpicture}
\begin{axis}[paperbars, width=\linewidth, height=4.1cm, bar width=9pt, ymax=1.2, ylabel={Success on the final task}, xticklabel style={text width=2cm},
  xmin=-0.5, xmax=2.5, xtick={0,1,2},
  xticklabels={{Own earlier\\task},{Hand-written,\\with read-back},{Hand-written,\\no read-back}}]
\addplot[armvalid] coordinates {(0,1.0000) +- (0,0.0000) (1,0.9888) +- (0,0.0044) (2,0.9333) +- (0,0.0090)};
\addplot[armood, nodes near coords, point meta=explicit symbolic, nodes near coords style={font=\tiny, text=badred, yshift=1pt}] coordinates {(0,0.4600) +- (0,0.0135) [0.46] (1,0.8833) +- (0,0.0134) [0.88] (2,0.5754) +- (0,0.0171) [0.58]};
\node[font=\tiny, text=slate] at (axis cs:0,1.12) {read-back 30\%};
\node[font=\tiny, text=slate] at (axis cs:1,1.12) {read-back 84\%};
\node[font=\tiny, text=slate] at (axis cs:2,1.12) {read-back 17\%};
\end{axis}
\end{tikzpicture}
\caption{Success when the agent did the earlier task itself and when we wrote it, with and without a read-back step.
Above each group: the share of runs with out-of-date experience that read back after the first write. Five changes,
six models, $\pm$ one standard error.}
\label{fig:authored}
\end{figure}
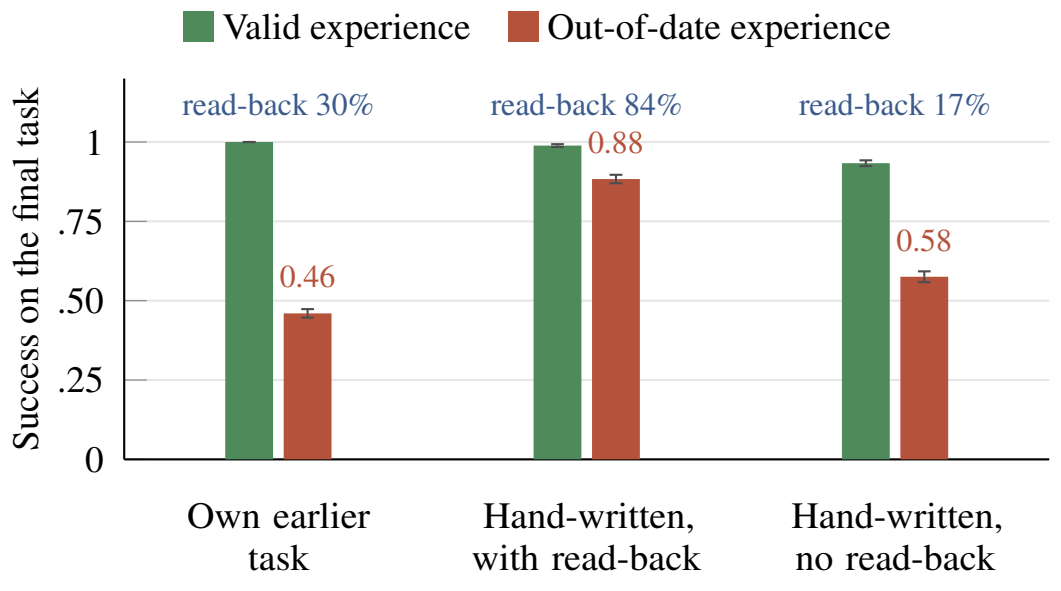

Studies of agent memory often give the agent histories written by the authors instead of letting it gain the
experience itself. We compare the two on the same tasks. The hand-written earlier task makes the correct call for its
version and receives the acknowledgement. In one form it then reads back the stored state and confirms it; in the
other it ends with ``Done. The event is scheduled.''

With a hand-written history that contains the read-back step, success with out-of-date experience is 88\%, against
46\% after the agent's own earlier task (Figure~\ref{fig:authored}). The agents copy the procedure of the history: 84\%
read back after their first write, against 30\% after their own earlier task. Without the read-back step, the read-back
rate falls to 17\% and success to 58\%. An evaluation with hand-written histories that contain a check step would
measure about a fifth of the cost that agents incur with their own experience (0.11 against 0.54). This may be one
reason why the failure has received little attention.

\section{Discussion}
\label{sec:discussion}

An agent's own successful experience changes how it uses a tool: it stops checking. After a silent change, earlier
successes become failures that the agent reports as successes, and with incomplete documentation the experienced agent
does worse than an agent without experience. Storing what worked is therefore not enough; a memory system also needs a
way to notice that a stored usage no longer holds. In our tests, checking the stored result did this reliably, at a
cost in model calls; version labels and pointers to the documentation did not. Evaluations of memory under change
should use the agent's own experience and incomplete documentation, because hand-written histories and complete
documentation both hide the problem (Appendix~\ref{app:discussion}).

% Required by ARR; does not count toward the page limit. Starts on a new page so that the eight pages of content end
% visibly before it.
\clearpage
\section*{Limitations}

The services are synthetic and small, and each version pair differs in one interface detail. Real changes often come
in groups, and some produce partial errors; we chose single silent changes to isolate the effect of experience. The
tasks are short, and the experience consists of one to ten earlier tasks of the same kind. With complete
documentation, the documentation is correct and always one call away; in practice documentation can lag behind the
service. With incomplete documentation, the earlier request states what the values mean, so the experience contains an
explicit rule; we did not test experience that the agent had to find out by trial. All six models run in the same
minimal agent loop; agent frameworks with their own planning or verification steps may
behave differently, as the effect of the instruction to check the result suggests. Most remedies were tested only with
the experience in the conversation.

Most excluded episodes come from three models whose earlier task on the weekday version in which 0 is Monday never
succeeded (Appendix~\ref{app:calibration}). Their weekday estimates therefore condition on runs in which the earlier
task succeeded. The standard errors resample task instances and treat the models and changes as fixed; they do not
cover variation across services or models beyond those tested and may be optimistic. The conditions with incomplete
documentation were not preregistered. With incomplete documentation, the success of agents without experience depends
on whether the running version matches what the agent assumes by default; both directions are run for every change.
The measures of reasoning text and of completion reports rely on keyword patterns (Appendices~\ref{app:reasoning}
and~\ref{app:reports}).

% Optional statements; to be written.
% \section*{Ethics Statement}

% Acknowledgments are omitted in the anonymous review version.
% \section*{Acknowledgments}


\bibliography{custom}

@misc{li2026world,
  title = {{The World Won't Stay Still: Programmable Evolution for Agent Benchmarks}},
  author = {Guangrui Li and Yaochen Xie and Yi Liu and Ziwei Dong and Xingyuan Pan and Tianqi Zheng and Jason Choi and Michael J. Morais and Binit Jha and Shaunak Mishra and Bingrou Zhou and Chen Luo and Monica Xiao Cheng and Dawn Song},
  year = {2026},
  eprint = {2603.05910},
  archivePrefix = {arXiv},
  howpublished = {arXiv:2603.05910}
}

@misc{nakayashiki2026stale,
  title = {{When Stale Constraints Go Unchecked: Budgeted Verification Failures in Inherited Agent Memory}},
  author = {Kazuki Nakayashiki},
  year = {2026},
  eprint = {2608.25553},
  archivePrefix = {arXiv},
  howpublished = {arXiv:2608.25553}
}

@misc{ye2026setshifting,
  title = {{Set-shifting Behavioral Test for Harnessed Agents}},
  author = {Ye, Ziwei},
  year = {2026},
  eprint = {2607.13396},
  archivePrefix = {arXiv},
  howpublished = {arXiv:2607.13396}
}

@misc{elhattami2026skillnb,
  title = {{SKILL.nb: Selective Formalization and Gated Execution for Durable Agent Workflows}},
  author = {El Hattami, Amine and Nicolas Chapados and Christopher Pal},
  year = {2026},
  eprint = {2606.08049},
  archivePrefix = {arXiv},
  howpublished = {arXiv:2606.08049}
}

@inproceedings{chen2025toolevo,
  title = {{Learning Evolving Tools for Large Language Models}},
  author = {Guoxin Chen and Zhong Zhang and Xin Cong and Fangda Guo and Yesai Wu and Yankai Lin and Wenzheng Feng and Yasheng Wang},
  booktitle = {International Conference on Learning Representations (ICLR)},
  year = {2025},
  note = {arXiv:2410.06617}
}

@misc{chen2026compliance,
  title = {{The Compliance Trap: Diagnosing How AI Agents Consume Conflicting Memory}},
  author = {Yixiong Chen and Xinyi Bai and Alan Yuille},
  year = {2026},
  eprint = {2607.10608},
  archivePrefix = {arXiv},
  howpublished = {arXiv:2607.10608}
}

@misc{chao2026stale,
  title = {{STALE: Can LLM Agents Know When Their Memories Are No Longer Valid?}},
  author = {Hanxiang Chao and Yihan Bai and Rui Sheng and Tianle Li and Yushi Sun},
  year = {2026},
  eprint = {2605.06527},
  archivePrefix = {arXiv},
  howpublished = {arXiv:2605.06527}
}

@misc{englander2026explore,
  title = {{Agents Explore but Agents Ignore: LLMs Lack Environmental Curiosity}},
  author = {Leon Engländer and Sophia Althammer and Ahmet Üstün and Matthias Gallé and Tom Sherborne},
  year = {2026},
  eprint = {2604.17609},
  archivePrefix = {arXiv},
  howpublished = {arXiv:2604.17609}
}

@misc{liu2026mcpevol,
  title = {{MCPEvol-Bench: Benchmarking LLM Agent Performance Across Dynamic Evolutions of MCP Servers}},
  author = {Huanxi Liu and Kun Hu and Jiaqi Liao and Qiang Wang and Pengfei Qian and YuanZhao Zhai and Dawei Feng and Bo Ding and Huaimin Wang},
  year = {2026},
  eprint = {2607.14642},
  archivePrefix = {arXiv},
  howpublished = {arXiv:2607.14642}
}

@misc{toh2026scramble,
  title = {{ScrambleToolBench: Agents Search Exhaustively Even When Their Own Map Points to the Next Step}},
  author = {Vernon Toh and Navonil Majumder and Zhengyuan Liu and Nancy F. Chen and Soujanya Poria},
  year = {2026},
  eprint = {2608.02358},
  archivePrefix = {arXiv},
  howpublished = {arXiv:2608.02358}
}

@misc{wang2026memtrap,
  title = {{MemTrapBench: Benchmarking Cognitive Traps in LLM Memory Use}},
  author = {Mengru Wang and Haozhe Luo and Zhenqian Xu and Zhixiang Cui and Haoming Xu and Qu Yang and Jizhan Fang and Junfeng Fang and Ningyu Zhang},
  year = {2026},
  eprint = {2608.20202},
  archivePrefix = {arXiv},
  howpublished = {arXiv:2608.20202}
}

@misc{fan2026statemem,
  title = {{Can Agent Memory Systems Track Evolving State?}},
  author = {Xinyi Fan and Miri Liu and Ruozhen Yang and Siru Ouyang and Jiawei Han},
  year = {2026},
  eprint = {2608.19652},
  archivePrefix = {arXiv},
  howpublished = {arXiv:2608.19652}
}

@misc{hu2026continual,
  title = {{When Continual Learning Moves to Memory: A Study of Experience Reuse in LLM Agents}},
  author = {Qisheng Hu and Quanyu Long and Wenya Wang},
  year = {2026},
  eprint = {2604.27003},
  archivePrefix = {arXiv},
  howpublished = {arXiv:2604.27003}
}

@inproceedings{xiong2025memory,
  title = {{How Memory Management Impacts LLM Agents: An Empirical Study of Experience-Following Behavior}},
  author = {Zidi Xiong and Yuping Lin and Wenya Xie and Pengfei He and Zirui Liu and Jiliang Tang and Himabindu Lakkaraju and Zhen Xiang},
  booktitle = {Proceedings of the Annual Meeting of the Association for Computational Linguistics (ACL)},
  year = {2026},
  note = {arXiv:2505.16067}
}

@misc{wan2026inertia,
  title = {{Mitigating Conversational Inertia in Multi-Turn Agents}},
  author = {Yang Wan and Zheng Cao and Zhenhao Zhang and Zhengwen Zeng and Shuheng Shen and Changhua Meng and Linchao Zhu},
  year = {2026},
  eprint = {2602.03664},
  archivePrefix = {arXiv},
  howpublished = {arXiv:2602.03664}
}

@misc{sinha2025illusion,
  title = {{The Illusion of Diminishing Returns: Measuring Long Horizon Execution in LLMs}},
  author = {Akshit Sinha and Arvindh Arun and Shashwat Goel and Steffen Staab and Jonas Geiping},
  year = {2025},
  eprint = {2509.09677},
  archivePrefix = {arXiv},
  howpublished = {arXiv:2509.09677}
}

@misc{laban2025lost,
  title = {{LLMs Get Lost In Multi-Turn Conversation}},
  author = {Philippe Laban and Hiroaki Hayashi and Yingbo Zhou and Jennifer Neville},
  year = {2025},
  eprint = {2505.06120},
  archivePrefix = {arXiv},
  howpublished = {arXiv:2505.06120}
}

@misc{cheng2025temporal,
  title = {{Your LLM Agents are Temporally Blind: The Misalignment Between Tool Use Decisions and Human Time Perception}},
  author = {Yize Cheng and Arshia Soltani Moakhar and Chenrui Fan and Parsa Hosseini and Kazem Faghih and Zahra Sodagar and Wenxiao Wang and Soheil Feizi},
  year = {2025},
  eprint = {2510.23853},
  archivePrefix = {arXiv},
  howpublished = {arXiv:2510.23853}
}

@inproceedings{sun2024toolsfail,
  title = {{Tools Fail: Detecting Silent Errors in Faulty Tools}},
  author = {Jimin Sun and So Yeon Min and Yingshan Chang and Yonatan Bisk},
  booktitle = {Proceedings of the Conference on Empirical Methods in Natural Language Processing (EMNLP)},
  year = {2024},
  note = {arXiv:2406.19228}
}

@misc{zhu2026toolmaze,
  title = {{When Tools Fail: Benchmarking Dynamic Replanning and Anomaly Recovery in LLM Agents}},
  author = {Dongsheng Zhu and Xuchen Ma and Yucheng Shen and Xiang Li and Yukun Zhao and Shuaiqiang Wang and Lingyong Yan and Dawei Yin},
  year = {2026},
  eprint = {2606.05806},
  archivePrefix = {arXiv},
  howpublished = {arXiv:2606.05806}
}

@misc{li2026silentprobe,
  title = {{SilentProbe: Measuring Silent Failure in Production APIs Used as Agent Tools}},
  author = {Zongrong Li and Shengkun Ye and Feiyou Guo and Zuoyou Dang},
  year = {2026},
  eprint = {2609.00035},
  archivePrefix = {arXiv},
  howpublished = {arXiv:2609.00035}
}

@misc{advani2026falsesuccess,
  title = {{From Confident Closing to Silent Failure: Characterizing False Success in LLM Agents}},
  author = {Laksh Advani},
  year = {2026},
  eprint = {2606.09863},
  archivePrefix = {arXiv},
  howpublished = {arXiv:2606.09863}
}

@misc{qian2026temporal,
  title = {{Toward User-Conditioned Evaluation of Personal LLM Agents under Temporal Interventions}},
  author = {Pin Qian and Su Wang and Yihang Chen and Qiaolin Yu and Xiaoyuan Wang and Zhitong Guo and Zhicheng Wang and Junxian You},
  year = {2026},
  eprint = {2607.21635},
  archivePrefix = {arXiv},
  howpublished = {arXiv:2607.21635}
}

@inproceedings{wang2025libevolution,
  title = {{LLMs Meet Library Evolution: Evaluating Deprecated API Usage in LLM-based Code Completion}},
  author = {Chong Wang and Kaifeng Huang and Jian Zhang and Yebo Feng and Lyuye Zhang and Yang Liu and Xin Peng},
  booktitle = {Proceedings of the IEEE/ACM International Conference on Software Engineering (ICSE)},
  year = {2025},
  note = {arXiv:2406.09834}
}

@misc{islah2024gitchameleon,
  title = {{GitChameleon: Unmasking the Version-Switching Capabilities of Code Generation Models}},
  author = {Nizar Islah and Justine Gehring and Diganta Misra and Eilif Muller and Irina Rish and Terry Yue Zhuo and Massimo Caccia},
  year = {2024},
  eprint = {2411.05830},
  archivePrefix = {arXiv},
  howpublished = {arXiv:2411.05830}
}

@misc{liu2024codeupdatearena,
  title = {{CodeUpdateArena: Benchmarking Knowledge Editing on API Updates}},
  author = {Zeyu Leo Liu and Shrey Pandit and Xi Ye and Eunsol Choi and Greg Durrett},
  year = {2024},
  eprint = {2407.06249},
  archivePrefix = {arXiv},
  howpublished = {arXiv:2407.06249}
}

@misc{packer2023memgpt,
  title = {{MemGPT: Towards LLMs as Operating Systems}},
  author = {Charles Packer and Sarah Wooders and Kevin Lin and Vivian Fang and Shishir G. Patil and Ion Stoica and Joseph E. Gonzalez},
  year = {2023},
  eprint = {2310.08560},
  archivePrefix = {arXiv},
  howpublished = {arXiv:2310.08560}
}

@inproceedings{zhao2024expel,
  title = {{ExpeL: LLM Agents Are Experiential Learners}},
  author = {Andrew Zhao and Daniel Huang and Quentin Xu and Matthieu Lin and Yong-Jin Liu and Gao Huang},
  booktitle = {Proceedings of the AAAI Conference on Artificial Intelligence},
  year = {2024},
  note = {arXiv:2308.10144}
}

@misc{wang2024awm,
  title = {{Agent Workflow Memory}},
  author = {Zora Zhiruo Wang and Jiayuan Mao and Daniel Fried and Graham Neubig},
  year = {2024},
  eprint = {2409.07429},
  archivePrefix = {arXiv},
  howpublished = {arXiv:2409.07429}
}

@misc{wang2023voyager,
  title = {{Voyager: An Open-Ended Embodied Agent with Large Language Models}},
  author = {Guanzhi Wang and Yuqi Xie and Yunfan Jiang and Ajay Mandlekar and Chaowei Xiao and Yuke Zhu and Linxi Fan and Anima Anandkumar},
  year = {2023},
  eprint = {2305.16291},
  archivePrefix = {arXiv},
  howpublished = {arXiv:2305.16291}
}

@inproceedings{shinn2023reflexion,
  title = {{Reflexion: Language Agents with Verbal Reinforcement Learning}},
  author = {Noah Shinn and Federico Cassano and Edward Berman and Ashwin Gopinath and Karthik Narasimhan and Shunyu Yao},
  booktitle = {Advances in Neural Information Processing Systems (NeurIPS)},
  year = {2023},
  note = {arXiv:2303.11366}
}

@misc{chhikara2025mem0,
  title = {{Mem0: Building Production-Ready AI Agents with Scalable Long-Term Memory}},
  author = {Prateek Chhikara and Dev Khant and Saket Aryan and Taranjeet Singh and Deshraj Yadav},
  year = {2025},
  eprint = {2504.19413},
  archivePrefix = {arXiv},
  howpublished = {arXiv:2504.19413}
}

@misc{yao2024taubench,
  title = {{$\tau$-bench: A Benchmark for Tool-Agent-User Interaction in Real-World Domains}},
  author = {Shunyu Yao and Noah Shinn and Pedram Razavi and Karthik Narasimhan},
  year = {2024},
  eprint = {2406.12045},
  archivePrefix = {arXiv},
  howpublished = {arXiv:2406.12045}
}

@misc{lu2024toolsandbox,
  title = {{ToolSandbox: A Stateful, Conversational, Interactive Evaluation Benchmark for LLM Tool Use Capabilities}},
  author = {Jiarui Lu and Thomas Holleis and Yizhe Zhang and Bernhard Aumayer and Feng Nan and Felix Bai and Shuang Ma and Shen Ma and Mengyu Li and Guoli Yin and Zirui Wang and Ruoming Pang},
  year = {2024},
  eprint = {2408.04682},
  archivePrefix = {arXiv},
  howpublished = {arXiv:2408.04682}
}

@inproceedings{trivedi2024appworld,
  title = {{AppWorld: A Controllable World of Apps and People for Benchmarking Interactive Coding Agents}},
  author = {Harsh Trivedi and Tushar Khot and Mareike Hartmann and Ruskin Manku and Vinty Dong and Edward Li and Shashank Gupta and Ashish Sabharwal and Niranjan Balasubramanian},
  booktitle = {Proceedings of the Annual Meeting of the Association for Computational Linguistics (ACL)},
  year = {2024},
  note = {arXiv:2407.18901}
}

@misc{wang2025pi,
  title = {{Unable to Forget: Proactive Interference Reveals Working Memory Limits in LLMs Beyond Context Length}},
  author = {Chupei Wang and Jiaqiu Vince Sun},
  year = {2025},
  eprint = {2506.08184},
  archivePrefix = {arXiv},
  howpublished = {arXiv:2506.08184}
}

@misc{du2026memory,
  title = {{Memory for Autonomous LLM Agents: Mechanisms, Evaluation, and Emerging Frontiers}},
  author = {Pengfei Du},
  year = {2026},
  eprint = {2603.07670},
  archivePrefix = {arXiv},
  howpublished = {arXiv:2603.07670}
}

@misc{zhou2026externalization,
  title = {{Externalization in LLM Agents: A Unified Review of Memory, Skills, Protocols and Harness Engineering}},
  author = {Chenyu Zhou and Huacan Chai and Wenteng Chen and Zihan Guo and Rong Shan and Yuanyi Song and Tianyi Xu and Yingxuan Yang and Aofan Yu and Weiming Zhang and Congming Zheng and Jiachen Zhu and Zeyu Zheng and Zhuosheng Zhang and Xingyu Lou and Changwang Zhang and Zhihui Fu and Jun Wang and Weiwen Liu and Jianghao Lin and Weinan Zhang},
  year = {2026},
  eprint = {2604.08224},
  archivePrefix = {arXiv},
  howpublished = {arXiv:2604.08224}
}

@misc{chen2026contextual,
  title = {{Breaking Contextual Inertia: Reinforcement Learning with Single-Turn Anchors for Stable Multi-Turn Interaction}},
  author = {Xingwu Chen and Zhanqiu Zhang and Yiwen Guo and Difan Zou},
  year = {2026},
  eprint = {2603.04783},
  archivePrefix = {arXiv},
  howpublished = {arXiv:2603.04783}
}

@misc{dabas2026tooldrift,
  title = {{Memory-Induced Tool-Drift in LLM Agents}},
  author = {Mahavir Dabas and Jihyun Jeong and Ming Jin and Ruoxi Jia},
  year = {2026},
  eprint = {2605.24941},
  archivePrefix = {arXiv},
  howpublished = {arXiv:2605.24941}
}

@misc{haq2025mentalset,
  title = {{Is your LLM trapped in a Mental Set? Investigative study on how mental sets affect the reasoning capabilities of LLMs}},
  author = {Saiful Haq and Niyati Chhaya and Piyush Pandey and Pushpak Bhattacharya},
  year = {2025},
  eprint = {2501.11833},
  archivePrefix = {arXiv},
  howpublished = {arXiv:2501.11833}
}

@article{kim2025inflexible,
  title = {{Limitations of Large Language Models in Clinical Problem-Solving Arising from Inflexible Reasoning}},
  author = {Jonathan Kim and Anna Podlasek and Kie Shidara and Feng Liu and Ahmed Alaa and Danilo Bernardo},
  journal = {Scientific Reports},
  volume = {15},
  pages = {39426},
  year = {2025}
}

@article{luchins1942,
  title = {{Mechanization in Problem Solving: The Effect of Einstellung}},
  author = {Abraham S. Luchins},
  journal = {Psychological Monographs},
  volume = {54},
  number = {6},
  year = {1942}
}

@article{bilalic2008,
  title = {{Why Good Thoughts Block Better Ones: The Mechanism of the Pernicious Einstellung (Set) Effect}},
  author = {Merim Bilali{\'c} and Peter McLeod and Fernand Gobet},
  journal = {Cognition},
  volume = {108},
  number = {3},
  pages = {652--661},
  year = {2008}
}

@misc{todoist2026api,
  title = {{Todoist API v1 Reference}},
  author = {{Doist}},
  year = {2026},
  howpublished = {\url{https://developer.todoist.com/api/v1/}},
  note = {Sections ``Sync / Tasks / Add a task'' and ``Create Task''. Accessed 2026-10-12}
}

@misc{greirson2025mcptodoist,
  title = {{mcp-todoist}: {MCP} Server for Todoist},
  author = {{greirson}},
  year = {2025},
  howpublished = {GitHub repository, \url{https://github.com/greirson/mcp-todoist}},
  note = {File src/utils/priority-mapper.ts, commit d92eebc}
}

@misc{arora2025todoistmcp,
  title = {{todoist-mcp}: {MCP} Server for Todoist},
  author = {{mehularora8}},
  year = {2025},
  howpublished = {GitHub repository, \url{https://github.com/mehularora8/todoist-mcp}},
  note = {File todoist\_mcp\_server/todoist.py, commit 9dbba17}
}

@misc{mcp2025tools,
  title = {{Model Context Protocol Specification (2025-06-18): Tools}},
  author = {{Model Context Protocol}},
  year = {2025},
  howpublished = {\url{https://modelcontextprotocol.io/specification/2025-06-18/server/tools}}
}

@misc{python2026datetime,
  title = {{datetime --- Basic date and time types}},
  author = {{Python Software Foundation}},
  year = {2026},
  howpublished = {\url{https://docs.python.org/3/library/datetime.html}}
}

@misc{mdn2026getday,
  title = {{Date.prototype.getDay()}},
  author = {{MDN Web Docs}},
  year = {2026},
  howpublished = {\url{https://developer.mozilla.org/en-US/docs/Web/JavaScript/Reference/Global_Objects/Date/getDay}}
}

@misc{opengroup2018crontab,
  title = {{crontab}},
  author = {{The Open Group}},
  year = {2018},
  howpublished = {The Open Group Base Specifications Issue 7, \url{https://pubs.opengroup.org/onlinepubs/9699919799/utilities/crontab.html}}
}

\appendix
\section{Further Discussion}
\label{app:discussion}

\paragraph{Memory systems and agent harnesses.}
Most memory designs store what worked and assume that it stays true. Our results show that stored experience also
changes how an agent works: it stops checking. A memory system therefore needs a way to find out that a stored usage
no longer holds. Labelling memories with the service version does little, because agents do not act on the label. Two
measures worked in our tests. A harness can attach the current documentation, or a short instruction to read it, to
the request itself rather than to the system prompt; this helps only when the documentation contains the answer, and
it adds model calls even when the experience is still valid. A harness can also make the agent read back the result
of each write; this works whether or not the documentation is complete, at more than twice the model calls. Where the
documentation may be incomplete, checking the result is the safer choice. We tested these measures with the experience
in the conversation.

\paragraph{Tool design.}
Changes that the tool definition shows were copied less often than changes that only the documentation describes (old
usage 69\% against 99\%). Meanings such as which end of a priority scale is highest or which day is day 0 can be put
into the schema, for example as enum names or field descriptions, so that an agent that does not read the
documentation still sees them. Write responses that return the stored object let many agents see and correct their
error without being asked. Rejecting unknown fields would turn a silent change into an error; we did not test it.

\paragraph{Evaluation.}
How the experience is produced changes what an evaluation measures. Hand-written histories carry procedures that the
agent copies, and one read-back step in such a history hides most of the cost. Evaluations of agent memory under change
should use the agent's own trajectories, or report the procedures that their hand-written histories contain. They
should also include tasks with incomplete documentation. When the documentation spells out everything, agents without
experience almost always succeed, so neither the value of experience nor the harm of out-of-date experience relative
to none becomes visible.

\paragraph{Mechanism.}
We have not established why agents stop checking. The data are consistent with experience-following
\citep{xiong2025memory}: the more similar earlier tasks the agent has done, the more closely its first write follows
them. Where text is placed also matters. The same instruction and the same documentation were acted on when they were
attached to the request, and had little or a negative effect in the system prompt or a tool description.

\section{Services, Tool Definitions and Documentation}
\label{app:services}

{\raggedright
\paragraph{Tools.} The calendar service has \code{calendar\_create\_event}, \code{calendar\_update\_event},
\code{calendar\_delete\_event}, \code{calendar\_get\_event} and \code{calendar\_list\_events}. The task service has
\code{tasks\_list\_tasks}, \code{tasks\_read\_task} and either \code{tasks\_update\_priority} or
\code{tasks\_set\_review\_day}. The mail service, used only for the reply-default change, has
\code{mail\_find\_thread}, \code{mail\_create\_reply\_draft}, \code{mail\_get\_draft}, \code{mail\_update\_draft} and
\code{mail\_send\_draft}. Every service has \code{docs\_search} (``Search the reference documentation for this
service.''), which returns the reference documentation of the running version. Read tools return the stored state in
neutral terms (for example \code{participants}, or a priority shown as Highest, High, Medium or Low), so a read-back
shows whether the request took effect without revealing the field names of either version.
\par}

\paragraph{Typed definitions.} Below is the create tool of the calendar service in version B of the field-name change;
version A is identical except that \code{guests} is \code{attendees}.
\begin{lstlisting}
{"name": "calendar_create_event", "description": "Create a calendar event.",
 "parameters": {"type": "object", "properties": {"payload": {"type": "object", "properties": {
   "title": {"type": "string", "description": "Event title."},
   "start_time": {"type": "string", "description": "ISO 8601 start time."},
   "duration_minutes": {"type": "integer", "description": "Length of the event in minutes."},
   "guests": {"type": "array", "items": {"type": "string"},
              "description": "Email addresses of the people taking part."}},
   "required": ["title", "start_time", "duration_minutes"]}},
 "required": ["payload"], "additionalProperties": false}}
\end{lstlisting}
In the structure change, the payload has \code{end\_time} (A) or \code{duration\_minutes} (B), and an event without a
length lasts 30 minutes. In the value change, \code{priority} has the enum \code{urgent, high, normal, low} (A) or
\code{p0, p1, p2, p3} (B). In the priority-direction change, \code{priority} is an integer from 1 to 4 in both
versions, and in the weekday change \code{day} is an integer from 0 to 6 in both versions; the descriptions are
``Priority level.'' and ``Day of the week.''. In the documentation-only case, the calendar write tools take an untyped
\code{payload} (``Request fields. Field names and values follow the service reference.'') and the priority tool an
untyped string.

\paragraph{Documentation.} The documentation of each version has a section per tool with a field table or a value
list and an example. The weekday change, version A:
\begin{lstlisting}
# Task reference
## tasks_set_review_day
Sets the day of the week on which an existing task is reviewed. Arguments: `task_id` and `day`.
`day` is an integer from 0 to 6: 0 = Monday, 1 = Tuesday, 2 = Wednesday, 3 = Thursday,
4 = Friday, 5 = Saturday, 6 = Sunday.
Example arguments:
{"task_id": "tsk_100", "day": 5}
## tasks_list_tasks
Lists tasks whose title contains the given text, with their review day.
## tasks_read_task
Returns the stored task, including its review day.
\end{lstlisting}
Version B reads ``0 = Sunday, 1 = Monday, \dots, 6 = Saturday''. The priority-direction documentation reads ``1 is the
highest and 4 is the lowest'' (A) or ``4 is the highest and 1 is the lowest'' (B).

\paragraph{Tasks.} An earlier and a final task of the field-name change: ``Schedule a 45-minute event titled ``Hiring
Debrief'' on March 27, 2026 at 10:00 AM America/Los\_Angeles time, with Noah (noah@example.com) and Alice
(alice@example.com) as participants.'' Priority tasks ask the agent to ``set it to the highest priority level that the
task service offers''; weekday tasks to ``set its weekly review day to Wednesday''. Each task instance draws titles,
times, people and days from fixed pools with a seed determined by the change, the direction and the repeat index.

\section{Prompts}
\label{app:prompts}

\begin{description}[leftmargin=0pt,itemsep=2pt,font=\normalfont\itshape]
\item[System prompt.] ``You are an agent that completes the user's requests by calling the available tools. Carry out
each request with the tools, then reply with a brief final message.''
\item[Memory request (after the earlier task).] ``Write a short note for yourself to save as memory.md. It will be
shown to you at the start of future sessions with this service. Record anything from this session that would help with
similar requests. Reply with the note text only.''
\item[Memory header.] ``Notes you saved in earlier sessions (memory.md):''; with a version tag: ``Notes you saved in
earlier sessions (memory.md), written when the service was at version 2.2:'' (2.3 for valid experience).
\item[Retrieved-memory header.] ``Notes from earlier sessions that may be relevant (retrieved from memory.md):''
\item[Documentation header.] ``Reference documentation for this service:'', followed by the documentation of the
running version.
\item[Version and notice.] Appended to each service tool description: ``Service version 2.2.'' in the earlier task and
``Service version 2.3. The service was updated on 2026-03-01; see the reference documentation for details.'' in the
final task.
\item[Instruction to read the documentation first.] ``Before your first tool call that changes data in a request,
check the reference documentation for the exact fields and values to use.''
\item[Same, in the task message.] ``Before your first tool call
that changes data in this request, check the reference documentation for the exact fields and values to use.''
\item[Instruction to verify and correct.] ``Before you reply that a request is complete, read back the stored state and
confirm that every requested detail is in place. If something does not match, check the reference documentation and
correct it.''
\item[Skill.] The system prompt adds: ``You also have a workspace whose files persist across sessions. Python scripts in
the workspace can call any of the service tools with \code{import service} and \code{service.call(tool\_name,
arguments)}, which returns the same result as calling the tool directly.'' The earlier request adds: ``Do this by
writing a reusable Python script in your workspace that you can run for similar requests, and run it.''
\item[Hand-written history without read-back.] The final reply is ``Done. The event is scheduled.'' (calendar) or
``Done. The task priority is set.'' (task service). In the weekday change the latter reply is also used, although that
task sets a review day.
\end{description}

\section{Conditions and Episode Counts}
\label{app:conditions}

\begin{table*}[t]
\centering
\small
\setlength{\tabcolsep}{4pt}
\resizebox{\textwidth}{!}{%
\begin{tabular}{llllr}
\toprule
Setting & Conditions & Changes & Arms & Episodes per model \\
\midrule
Complete documentation (\S\ref{sec:effect}) & untyped tool; typed tool listing the fields; meaning of a number & 3 + 3 + 3 & none, valid, out-of-date & 324 \\
Incomplete documentation (\S\ref{sec:effect})$^\ddagger$ & earlier request states the meaning & 4 & none, valid, out-of-date & 288 \\
Where the experience is stored (\S\ref{sec:memory}) & conversation, memory file, retrieved memory, skill & 5 memory-needed & none (1 instance), valid, out-of-date & 280 \\
How much experience (\S\ref{sec:memory}) & 3 or 10 earlier tasks in the conversation & 5 typed & valid, out-of-date & 120 \\
Additions to the final task (\S\ref{sec:remedies}) & documentation in tool description / system prompt / task message; & 5 typed & out-of-date & 360 \\
  & version and notice; instruction to read first; instruction to verify & & & \\
  & nothing added (rerun); instruction in the task message; write returns the stored object$^\dagger$ & 5 typed & out-of-date; valid for stored object & 240 \\
  & instruction in the task message$^\ddagger$ & 5 typed & valid & 60 \\
  & memory file with version tag; memory file with current documentation & 5 memory-needed & valid, out-of-date & 120 \\
  & incomplete documentation; instruction to read the docs / to verify$^\ddagger$ & 4 & out-of-date & 192 \\
Who wrote the earlier task (\S\ref{sec:authored}) & hand-written earlier task with / without read-back step & 5 typed & valid, out-of-date & 120 \\
\midrule
 & & & Total per model & 2,104 \\
\bottomrule
\end{tabular}}
\caption{Conditions. Each change runs in both directions. Instances per change, direction and arm: six with complete
documentation and for the additions to the final task, three for the storage, amount and authorship conditions, and
twelve with incomplete documentation. The meaning-of-a-number case with complete documentation includes the
reply-default change, which is reported separately (Appendix~\ref{app:variant}). The nothing-added and valid arms of
the additions, and the single earlier task in the amount-of-experience comparison, come from the complete-documentation
conditions. $^\dagger$Run after the other conditions on the same task instances, with its own preregistration
(Appendix~\ref{app:followup}). $^\ddagger$Not preregistered. Six models, 12,624 episodes.}
\label{tab:conditions}
\end{table*}

Table~\ref{tab:conditions} lists the conditions. When the experience is stored in different forms, the five changes
are the field-name, structure, value, priority-direction and weekday changes, with final requests that refer to the
earlier task. With three or ten earlier tasks, the additional earlier tasks have titles that do not occur in the main
task pools, so no two tasks in an episode share a title. The step limit is 16 model calls for each earlier and final task, and the memory request is retried once if the
note is empty.

\section{Models}
\label{app:models}

We use GLM-5.3-Flash, Muse-Glimmer-30B, DeepSeek-V4-Flash (version 0731), MiniMax-M3, Qwen3.8-Flash and Inkling. All
models are used with native tool calling, temperature 0.7 and at most 4,096 output tokens per call. Requests that
failed were retried; an episode with a request that still failed is excluded.

\section{Preregistered Predictions}
\label{app:prereg}

\begin{table*}[t]
\centering
\small
\setlength{\tabcolsep}{4pt}
\resizebox{\textwidth}{!}{%
\begin{tabular}{lllcl}
\toprule
 & Contrast & Prediction & Estimate & Result \\
\midrule
% Source inlined from tables/rot_contrasts.tex
P1 & Cost, interface only in the documentation & $>0$ & $0.74 \pm 0.01$ & holds \\
P1 & Cost, change visible in the tool definition & $>0$ & $0.51 \pm 0.02$ & holds \\
P1 & Cost, change not visible in the tool definition & $>0$ & $0.58 \pm 0.02$ & holds \\
\midrule
P2 & Cost, visible change $-$ documentation only & none & $-0.23 \pm 0.02$ & negative \\
\midrule
P3 & Documentation in the task message & $>0$ & $0.53 \pm 0.01$ & holds \\
P3 & Documentation in the tool description & $>0$ & $0.09 \pm 0.02$ & holds \\
P3 & Documentation in the system prompt & none & $0.08 \pm 0.02$ & positive \\
P3 & Version number and update notice & none & $0.04 \pm 0.02$ & positive \\
P3 & Instruction: read the documentation first & none & $-0.17 \pm 0.02$ & negative \\
P3 & Instruction: verify and correct & $>0$ & $0.45 \pm 0.02$ & holds \\
\midrule
P4 & Success without experience, memory-needed tasks & $<0.20$ & $0.07 \pm 0.03$ & holds \\
P4 & Cost, earlier conversation & $>0$ & $0.53 \pm 0.02$ & holds \\
P4 & Cost, memory file & $>0$ & $0.27 \pm 0.02$ & holds \\
P4 & Cost, retrieved memory & $>0$ & $0.20 \pm 0.02$ & holds \\
P4 & Cost, skill & $>0$ & $0.30 \pm 0.03$ & holds \\
P4 & Cost, memory file with current documentation $-$ memory file & $<0$ & $-0.17 \pm 0.03$ & holds \\
P4 & Cost, memory file with version tag $-$ memory file & none & $-0.07 \pm 0.03$ & negative \\
\midrule
P5 & Cost, hand-written without $-$ with read-back step & $>0$ & $0.25 \pm 0.04$ & holds \\
P5 & Cost, own earlier task $-$ hand-written with read-back step & $>0$ & $0.43 \pm 0.02$ & holds \\
\midrule
P6 & Cost, three earlier tasks $-$ one & $>0$ & $0.12 \pm 0.03$ & holds \\
P6 & Cost, ten earlier tasks $-$ three & none & $0.05 \pm 0.02$ & positive \\
% End of tables/rot_contrasts.tex
\bottomrule
\end{tabular}}
\caption{Preregistered contrasts (estimate $\pm$ one standard error). P1--P2: complete documentation; P3: additions to the final
task, success with out-of-date experience under each addition minus success without it; P4: where the experience is
stored; P5: who wrote the earlier task; P6: how much experience. A prediction holds when the 95\% percentile interval lies on
the predicted side; for contrasts without a prediction, the sign is reported when the interval excludes zero.}
\label{tab:prereg}
\end{table*}

The preregistration fixed the models, the conditions, the number of instances, the seeds, the exclusion rules, the
predictions and the analysis script, and recorded the hashes of the code before the run. Table~\ref{tab:prereg} lists
all preregistered contrasts. Every directional prediction held. The bootstrap resamples task instances (change,
direction, repeat index) within strata of change and direction. Instances with the same change share seeds across
conditions, so contrasts between conditions resample the same instances jointly.

\section{Additional Results}

\subsection{Main results in numbers}
\label{app:main}

\begin{table*}[!htbp]
\centering
\small
\setlength{\tabcolsep}{4pt}
\resizebox{\textwidth}{!}{%
\begin{tabular}{lcccccccc}
\toprule
 & \multicolumn{4}{c}{Success} & Lookup before & Old usage in & Read-back after & Failures reported \\
\cmidrule(lr){2-5}
Change & No experience & Valid & Out-of-date & Cost & first write & first write & first write & as done \\
\midrule
\multicolumn{9}{l}{\emph{Incomplete documentation (four changes)}} \\
% Source inlined from tables/ambig_effect.tex
Meaning of a number & $0.63 \pm 0.01$ & $1.00 \pm 0.00$ & $0.23 \pm 0.01$ & $0.77 \pm 0.01$ & 0.35 / 0.01 & 1.00 & 0.30 & 0.92 \\
% End of tables/ambig_effect.tex
\midrule
\multicolumn{9}{l}{\emph{Complete documentation}} \\
% Source inlined from tables/rot_main.tex
Field or value, untyped tool & $1.00 \pm 0.00$ & $1.00 \pm 0.00$ & $0.26 \pm 0.01$ & $0.74 \pm 0.01$ & 1.00 / 0.01 & 0.99 & 0.27 & 1.00 \\
Field or value, in the definition & $1.00 \pm 0.00$ & $1.00 \pm 0.00$ & $0.49 \pm 0.02$ & $0.51 \pm 0.02$ & 0.25 / 0.06 & 0.69 & 0.22 & 1.00 \\
Meaning of a number & $0.88 \pm 0.01$ & $1.00 \pm 0.00$ & $0.42 \pm 0.02$ & $0.58 \pm 0.02$ & 0.70 / 0.00 & 0.99 & 0.43 & 0.99 \\
% End of tables/rot_main.tex
\bottomrule
\end{tabular}}
\caption{Success on the final task with no, valid and out-of-date experience, and the cost (valid minus out-of-date),
each $\pm$ one standard error. Lookup is given for no experience / out-of-date experience; old usage, read-back and
the share of failed runs whose final reply reports the task as done are given for out-of-date experience. ``Field or
value'': the field-name, structure and value changes, with an untyped write tool or with a typed definition that lists
the current fields. ``Meaning of a number'': the priority-direction and weekday changes, and with incomplete
documentation also the month and duration changes. Six models.}
\label{tab:main}
\end{table*}

\begin{table*}[!htbp]
\centering
\small
\setlength{\tabcolsep}{4pt}
\resizebox{\textwidth}{!}{%
\begin{tabular}{lccccccc}
\toprule
 & \multicolumn{4}{c}{Success} & Lookup before & Old usage in & Content right in \\
\cmidrule(lr){2-5}
Where the experience is stored & No experience & Valid & Out-of-date & Cost & first write & first write & first write \\
\midrule
% Source inlined from tables/rot_memory.tex
Conversation & $0.07 \pm 0.03$ & $1.00 \pm 0.00$ & $0.47 \pm 0.02$ & $0.53 \pm 0.02$ & 0.02 & 0.83 & 1.00 \\
Memory file & $0.07 \pm 0.03$ & $0.99 \pm 0.01$ & $0.72 \pm 0.03$ & $0.27 \pm 0.02$ & 0.24 & 0.37 & 0.99 \\
Retrieved memory (top 3) & $0.10 \pm 0.04$ & $0.96 \pm 0.01$ & $0.76 \pm 0.01$ & $0.20 \pm 0.02$ & 0.34 & 0.27 & 0.98 \\
Skill (script and note) & $0.03 \pm 0.02$ & $0.91 \pm 0.01$ & $0.61 \pm 0.02$ & $0.30 \pm 0.03$ & 0.27 & 0.71 & 0.95 \\
\midrule
Memory file + version tag & -- & $0.97 \pm 0.01$ & $0.77 \pm 0.01$ & $0.20 \pm 0.02$ & 0.44 & 0.29 & 1.00 \\
Memory file + current docs & -- & $0.99 \pm 0.00$ & $0.89 \pm 0.01$ & $0.10 \pm 0.01$ & 0.37 & 0.18 & 1.00 \\
% End of tables/rot_memory.tex
\bottomrule
\end{tabular}}
\caption{Experience stored in different forms, with final requests that refer to the earlier task. Success and cost
$\pm$ one standard error; lookup, old usage and content of the first write for out-of-date experience. ``Content right'' means that the first write
carries the information from the earlier task (for example the right participants), whatever the field names. Five
changes, six models. The last two rows add the service version or the current documentation to the memory file
(Section~\ref{sec:remedies}); they were run with valid and out-of-date experience only.}
\label{tab:memory}
\end{table*}

Table~\ref{tab:main} gives the numbers behind Figures~\ref{fig:effect} and~\ref{fig:checking}a, and
Table~\ref{tab:memory} those behind Figure~\ref{fig:memory}a. The lower cost of retrieved memory is not due to missing
information: the retrieved entries contain the needed information in 90\% of runs, and in those runs the cost is 0.18.
Without experience, agents with complete documentation fail mainly on the weekday change (78\% success), where all 16
failures use the other numbering without consulting the documentation (Appendix~\ref{app:variant}).

\subsection{Additions to the final task}
\label{app:context}

\begin{table*}[t]
\centering
\small
\setlength{\tabcolsep}{4pt}
\resizebox{\textwidth}{!}{%
\begin{tabular}{lccccccc}
\toprule
 & \multicolumn{3}{c}{Success, out-of-date} & Lookup before & Old usage in & Read-back after & Model \\
\cmidrule(lr){2-4}
Added to the final task & All & Field or value & Meaning & first write & first write & first write & calls \\
\midrule
% Source inlined from tables/rot_context.tex
Nothing added & $0.46 \pm 0.01$ & $0.49 \pm 0.02$ & $0.42 \pm 0.02$ & 0.04 & 0.80 & 0.30 & 4.0 \\
Documentation in the task message & $0.99 \pm 0.00$ & $0.99 \pm 0.01$ & $0.99 \pm 0.01$ & 0.09 & 0.01 & 0.30 & 2.9 \\
Documentation in the tool description & $0.55 \pm 0.01$ & $0.61 \pm 0.02$ & $0.47 \pm 0.02$ & 0.08 & 0.62 & 0.30 & 3.6 \\
Documentation in the system prompt & $0.54 \pm 0.01$ & $0.57 \pm 0.02$ & $0.48 \pm 0.02$ & 0.07 & 0.63 & 0.29 & 3.5 \\
Version number and update notice & $0.50 \pm 0.01$ & $0.55 \pm 0.02$ & $0.44 \pm 0.02$ & 0.07 & 0.76 & 0.30 & 3.9 \\
Instruction: read the documentation first & $0.29 \pm 0.01$ & $0.26 \pm 0.01$ & $0.35 \pm 0.02$ & 0.04 & 0.96 & 0.29 & 4.0 \\
Instruction: verify and correct & $0.91 \pm 0.01$ & $0.93 \pm 0.01$ & $0.87 \pm 0.03$ & 0.03 & 0.84 & 0.92 & 6.5 \\
% End of tables/rot_context.tex
\bottomrule
\end{tabular}}
\caption{Additions to the final task, complete documentation, out-of-date experience. ``Nothing added'' is the
out-of-date arm without additions (five changes). With
valid experience the same changes give 1.00 success and 2.9 model calls.}
\label{tab:context}
\end{table*}

Table~\ref{tab:context} gives the full results. In the earlier task, the share of agents who search the
documentation is 98\% under the instruction to read it first and 46--64\% under the other conditions.

\subsection{Who wrote the earlier task, and how much experience}

\begin{table}[!htbp]
\centering
\footnotesize
\setlength{\tabcolsep}{3pt}
\begin{adjustbox}{max width=\linewidth}
\begin{tabular}{lccc}
\toprule
Earlier task(s) & Valid & Out-of-date & Cost \\
\midrule
% Source inlined from tables/rot_authored_dose.tex
Own earlier task & $1.00 \pm 0.00$ & $0.46 \pm 0.01$ & $0.54 \pm 0.01$ \\
Hand-written, with read-back & $0.99 \pm 0.00$ & $0.88 \pm 0.01$ & $0.11 \pm 0.01$ \\
Hand-written, read-back removed & $0.93 \pm 0.01$ & $0.58 \pm 0.02$ & $0.36 \pm 0.02$ \\
\midrule
One earlier task & $1.00 \pm 0.00$ & $0.46 \pm 0.01$ & $0.54 \pm 0.01$ \\
Three earlier tasks & $0.99 \pm 0.01$ & $0.33 \pm 0.02$ & $0.66 \pm 0.02$ \\
Ten earlier tasks & $0.97 \pm 0.01$ & $0.26 \pm 0.02$ & $0.71 \pm 0.02$ \\
% End of tables/rot_authored_dose.tex
\bottomrule
\end{tabular}
\end{adjustbox}
\caption{Who wrote the earlier task (top) and how many earlier tasks there are (bottom): success and cost $\pm$ one
standard error. The first row of each block is the agent's own single earlier task.}
\label{tab:authored-dose}
\end{table}

Table~\ref{tab:authored-dose} gives the numbers behind Figures~\ref{fig:memory}b and~\ref{fig:authored}.

\subsection{Per model}

\begin{table}[!htbp]
\centering
\footnotesize
\setlength{\tabcolsep}{3pt}
\begin{adjustbox}{max width=\linewidth}
\begin{tabular}{lccccccc}
\toprule
 & \multicolumn{3}{c}{Success} & Old & Read- & \multicolumn{2}{c}{Added to final task} \\
\cmidrule(lr){2-4}\cmidrule(lr){7-8}
Model & None & Valid & Out-of-date & usage & back & Docs in task & Verify \\
\midrule
% Source inlined from tables/rot_permodel.tex
GLM & 0.98 & 1.00 & 0.70 & 0.80 & 0.53 & 1.00 & 0.98 \\
Muse & 1.00 & 1.00 & 0.25 & 0.92 & 0.23 & 0.97 & 0.82 \\
DeepSeek & 0.90 & 1.00 & 0.35 & 0.61 & 0.00 & 1.00 & 0.83 \\
MiniMax & 0.92 & 1.00 & 0.05 & 0.95 & 0.00 & 0.98 & 0.81 \\
Qwen3.8 & 1.00 & 1.00 & 1.00 & 0.88 & 1.00 & 1.00 & 1.00 \\
Inkling & 0.88 & 1.00 & 0.38 & 0.64 & 0.00 & 1.00 & 0.98 \\
% End of tables/rot_permodel.tex
\bottomrule
\end{tabular}
\end{adjustbox}
\caption{Per model, complete documentation and typed tool definitions: success by arm, and old usage and read-back
with out-of-date experience; success with out-of-date experience when the documentation is added to the task message,
and under the instruction to verify and correct.}
\label{tab:permodel}
\end{table}

Table~\ref{tab:permodel} gives per-model results. Every model succeeds with valid experience and fails more often with
out-of-date experience, except Qwen3.8-Flash, which reads back after every write and corrects its errors. Every model
succeeds in at least 97\% of runs when the documentation is in the task message.

\subsection{Per change}
\label{app:variant}

\begin{table}[!htbp]
\centering
\footnotesize
\setlength{\tabcolsep}{3pt}
\begin{adjustbox}{max width=\linewidth}
\begin{tabular}{lccccc}
\toprule
 & \multicolumn{3}{c}{Success} & Lookup & Old usage \\
\cmidrule(lr){2-4}
Change & None & Valid & Out-of-date & (none) & (out-of-date) \\
\midrule
% Source inlined from tables/rot_variant.tex
Field name (docs only) & 1.00 & 1.00 & 0.24 & 1.00 & 1.00 \\
Structure (docs only) & 1.00 & 1.00 & 0.21 & 1.00 & 1.00 \\
Value (docs only) & 1.00 & 1.00 & 0.33 & 1.00 & 0.97 \\
\midrule
Field name & 0.99 & 1.00 & 0.49 & 0.07 & 0.65 \\
Structure & 1.00 & 1.00 & 0.46 & 0.07 & 0.61 \\
Value & 1.00 & 1.00 & 0.51 & 0.62 & 0.79 \\
\midrule
Priority direction & 0.97 & 1.00 & 0.36 & 0.92 & 1.00 \\
Weekday numbering & 0.78 & 1.00 & 0.48 & 0.49 & 0.98 \\
Reply default & 0.93 & 1.00 & 1.00 & 0.08 & 0.00 \\
% End of tables/rot_variant.tex
\bottomrule
\end{tabular}
\end{adjustbox}
\caption{Complete documentation by change. Top block: untyped tool; middle: typed tool that lists the current fields
or values; bottom: meaning of a number.}
\label{tab:variant}
\end{table}

Table~\ref{tab:variant} gives the results with complete documentation by change. When the typed definition lists the
current fields, agents without experience rarely search the documentation for the two calendar changes (7\%), whose typed definitions are self-explanatory, but often for the
value change (62\%). The reply-default change in the mail service (the reply goes to the sender only or to all
participants when the reply scope is omitted) shows no cost: with out-of-date experience the agents set the scope
explicitly and succeed in every run. As preregistered, it is not pooled with the other changes of meaning.

\subsection{Copying details}
\label{app:copy}

\begin{table}[!htbp]
\centering
\footnotesize
\setlength{\tabcolsep}{3pt}
\begin{adjustbox}{max width=\linewidth}
\begin{tabular}{lcccccc}
\toprule
 & \multicolumn{3}{c}{Same time format} & \multicolumn{3}{c}{Same field order} \\
\cmidrule(lr){2-4}\cmidrule(lr){5-7}
Case & Valid & Out-of-date & None & Valid & Out-of-date & None \\
\midrule
% Source inlined from tables/rot_copy.tex
Untyped tool & 1.00 & 1.00 & 0.77 & 0.76 & 0.76 & 0.82 \\
Typed tool & 1.00 & 1.00 & 0.99 & 0.83 & 0.64 & 0.74 \\
% End of tables/rot_copy.tex
\bottomrule
\end{tabular}
\end{adjustbox}
\caption{Calendar changes, complete documentation: share of first writes whose start time has the same format (seconds,
offset) and whose payload has the same field order as the earlier call. For agents without experience the comparison
is with the earlier call of the out-of-date run on the same task instance.}
\label{tab:copy}
\end{table}

Table~\ref{tab:copy} compares the first write with the earlier call. When the interface is described only in the
documentation, experienced agents reproduce the format of the start time in every run, agents without experience in
77\%. With typed definitions, nearly all agents use the same format, so this measure does not separate the arms there.
Field order does not separate the arms in either case.

\subsection{Reasoning text}
\label{app:reasoning}

\begin{table}[!htbp]
\centering
\footnotesize
\setlength{\tabcolsep}{3pt}
\begin{adjustbox}{max width=\linewidth}
\begin{tabular}{lccc}
\toprule
Arm & With text & Plans lookup & Refers to earlier \\
\midrule
% Source inlined from tables/rot_reasoning.tex
No experience & 517 / 574 & 0.69 & 0.02 \\
Valid experience & 390 / 566 & 0.06 & 0.24 \\
Out-of-date experience & 407 / 565 & 0.09 & 0.25 \\
% End of tables/rot_reasoning.tex
\bottomrule
\end{tabular}
\end{adjustbox}
\caption{Complete documentation, all changes except the reply default: runs with reasoning or reply text before the first write,
and the share of those that mention a plan to check the documentation or refer to an earlier task.}
\label{tab:reasoning}
\end{table}

We concatenate the reasoning and reply text of all turns before the first write. A run \emph{plans a lookup} if the
text contains a verb such as check, look up, consult, search, read, review, verify or confirm followed within 40
characters by docs, documentation, reference, API, schema or field name, or mentions \code{docs\_search}. It
\emph{refers to an earlier task} if it contains already, previous(ly), earlier, before, last time or same followed
within 50 characters by docs, documentation, reference, field, format, structure, payload, call or API, or one of the
phrases ``as before'', ``same as before'', ``like before'' and ``I already know''. Table~\ref{tab:reasoning} gives the
rates. Further examples from failed runs with out-of-date experience: ``Same pattern as before - find the task and
update priority to p0.'' (MiniMax-M3); ``I already know from the docs that priority 1 is the highest.'' (Inkling);
``I'll create this event using the same field structure.'' (DeepSeek-V4-Flash).

\subsection{Paths to success and failure}
\label{app:paths}

\begin{table}[!htbp]
\centering
\footnotesize
\setlength{\tabcolsep}{3pt}
\begin{adjustbox}{max width=\linewidth}
\begin{tabular}{lcc}
\toprule
Path & Runs & Share \\
\midrule
% Source inlined from tables/rot_paths.tex
no lookup, old usage, no read-back, failure & 183 & 0.52 \\
no lookup, old usage, read back, success & 94 & 0.27 \\
no lookup, current usage, no read-back, success & 45 & 0.13 \\
looked up, current usage, no read-back, success & 12 & 0.03 \\
no lookup, current usage, read back, success & 7 & 0.02 \\
no lookup, current usage, no read-back, failure & 3 & 0.01 \\
% End of tables/rot_paths.tex
\bottomrule
\end{tabular}
\end{adjustbox}
\caption{Complete documentation, typed tool definitions, out-of-date experience (350 runs): the six most frequent combinations of lookup
before the first write, old usage in the first write, read-back after it, and outcome.}
\label{tab:paths}
\end{table}

Table~\ref{tab:paths} lists the most frequent paths. Of the 184 runs that wrote the old usage and did not read back,
183 failed; of the 97 runs that wrote the old usage and read back, 94 succeeded.

\subsection{Completion reports}
\label{app:reports}

We classify the final reply of a run as reporting the task as done if it contains one of the words scheduled, created,
set, updated, done, completed, successfully, moved, changed, added or saved, or the phrases ``has been'', ``is now'' or
``now shows'', and none of unable, could not, cannot, failed, unfortunately, sorry, error, did not or ``was not''. Runs
that stop at the step limit have no final reply and count as not reporting. With complete documentation, 347 of the 348
failed runs with out-of-date experience end with such a reply. With incomplete documentation, 405 of 438 do; 12 tell
the user that the write did not take effect, 13 stop at the step limit, and 8 end with a reply that matches neither
pattern.

\section{Instruction in the Task Message and Stored-Object Responses}
\label{app:followup}

These conditions were run after the others, with their own preregistration (models, conditions, instances,
predictions, analysis script and code hashes). They use the six models, the five typed changes, both directions and six
task instances per cell, with the same seeds as before, so the task instances equal those of the other additions to the
final task. There are three conditions, all with the earlier task live in the same conversation:
\begin{description}[leftmargin=0pt,itemsep=2pt,font=\normalfont\itshape]
\item[Nothing added.] The out-of-date condition with complete documentation, run again.
\item[Instruction in the task message.] The final task message ends with ``Before your first tool call that changes
data in this request, check the reference documentation for the exact fields and values to use.'' The earlier task
and the system prompt are unchanged.
\item[Write returns the stored object.] In both phases, every accepted write response also contains \code{stored}, the
object as the read tools return it, for example \code{\{"accepted": true, "event\_id": "evt\_600", "stored":
\{\dots, "participants": []\}\}}. Run with valid and out-of-date experience.
\end{description}
The run has 1,440 episodes; 11 are excluded because the earlier task never succeeded (Inkling 7, DeepSeek-V4-Flash 3,
MiniMax-M3 1, all on the weekday change), and none for failed model requests. The preregistered prediction (higher success with
out-of-date experience when writes return the stored object) held: $+0.34 \pm 0.02$. The instruction in the task message
had no directional prediction; its difference from the plain condition is $+0.54 \pm 0.02$. The cost under write
responses that return the stored object is $0.21 \pm 0.01$. The rerun of the nothing-added condition differs from the original by
$-0.01 \pm 0.02$. Table~\ref{tab:followup} gives the process measures, and
Tables~\ref{tab:followup-full} and~\ref{tab:followup-model} give the results by kind of change and by model.

\begin{table}[!htbp]
\centering
\footnotesize
\setlength{\tabcolsep}{2.5pt}
\begin{adjustbox}{max width=\linewidth}
\begin{tabular}{@{}>{\raggedright\arraybackslash}p{2.35cm}cccccc@{}}
\toprule
 & & & Old & Read- & Second & \\
Condition & Success & Lookup & usage & back & write & Calls \\
\midrule
% Source inlined from tables/abl_main.tex
Nothing added & $0.45 \pm 0.02$ & 0.03 & 0.81 & 0.30 & 0.27 & 4.0 \\
Instruction in the task message & $0.99 \pm 0.00$ & 1.00 & 0.02 & 0.31 & 0.02 & 3.7 \\
Instruction in the system prompt$^*$ & $0.29 \pm 0.01$ & 0.04 & 0.96 & 0.29 & 0.26 & 4.0 \\
Write returns the stored object & $0.79 \pm 0.01$ & 0.02 & 0.83 & 0.48 & 0.70 & 4.6 \\
Write returns the stored object, valid experience & $1.00 \pm 0.00$ & 0.01 & 0.00 & 0.11 & 0.00 & 2.7 \\
% End of tables/abl_main.tex
\bottomrule
\end{tabular}
\end{adjustbox}
\caption{Instruction in the task message and stored-object responses, out-of-date experience unless marked (five changes, six models). Success $\pm$ one standard
error; lookup before, old usage in, and read-back after the first write; share of runs with a second write; model
calls. $^*$From Table~\ref{tab:context}, same task instances.}
\label{tab:followup}
\end{table}

\begin{table}[!htbp]
\centering
\footnotesize
\setlength{\tabcolsep}{3pt}
\begin{adjustbox}{max width=\linewidth}
\begin{tabular}{@{}>{\raggedright\arraybackslash}p{2.6cm}cccc@{}}
\toprule
Condition & All & Field or value & Meaning & Runs \\
\midrule
% Source inlined from tables/abl_full.tex
Nothing added & $0.45 \pm 0.02$ & $0.48 \pm 0.02$ & $0.39 \pm 0.02$ & 353 \\
Instruction in the task message & $0.99 \pm 0.00$ & $0.99 \pm 0.01$ & $0.99 \pm 0.01$ & 356 \\
Instruction in the system prompt$^*$ & $0.29 \pm 0.01$ & $0.26 \pm 0.01$ & $0.35 \pm 0.02$ & 360 \\
Write returns the stored object & $0.79 \pm 0.01$ & $0.81 \pm 0.02$ & $0.74 \pm 0.02$ & 360 \\
Write returns the stored object, valid experience & $1.00 \pm 0.00$ & $1.00 \pm 0.00$ & $1.00 \pm 0.00$ & 360 \\
% End of tables/abl_full.tex
\bottomrule
\end{tabular}
\end{adjustbox}
\caption{Instruction in the task message and stored-object responses: success $\pm$ one standard error, overall
and for the changes of a field or value (three) and of meaning (two). $^*$From Table~\ref{tab:context}.}
\label{tab:followup-full}
\end{table}

\begin{table}[!htbp]
\centering
\footnotesize
\setlength{\tabcolsep}{3pt}
\begin{adjustbox}{max width=\linewidth}
\begin{tabular}{lccccc}
\toprule
 & Nothing & \multicolumn{2}{c}{Instruction in} & \multicolumn{2}{c}{Write returns stored} \\
\cmidrule(lr){3-4}\cmidrule(lr){5-6}
Model & added & task msg. & system$^*$ & out-of-date & valid \\
\midrule
% Source inlined from tables/abl_permodel.tex
GLM & 0.62 & 1.00 & 0.58 & 0.97 & 1.00 \\
Muse & 0.30 & 1.00 & 0.13 & 0.57 & 1.00 \\
DeepSeek & 0.33 & 0.98 & 0.02 & 0.98 & 1.00 \\
MiniMax & 0.07 & 0.98 & 0.00 & 0.43 & 1.00 \\
Qwen3.8 & 1.00 & 0.98 & 1.00 & 1.00 & 1.00 \\
Inkling & 0.36 & 1.00 & 0.03 & 0.77 & 1.00 \\
% End of tables/abl_permodel.tex
\bottomrule
\end{tabular}
\end{adjustbox}
\caption{Instruction in the task message and stored-object responses: success by model. $^*$From
Table~\ref{tab:context}.}
\label{tab:followup-model}
\end{table}

\section{Incomplete Documentation}
\label{app:undocumented}

\begin{table*}[t]
\centering
\small
\setlength{\tabcolsep}{4pt}
\begin{tabular}{lcccccc}
\toprule
 & & Lookup before & Old usage in & Read-back after & Second & Model \\
Condition & Success & first write & first write & first write & write & calls \\
\midrule
% Source inlined from tables/ambig_main.tex
No experience & $0.63 \pm 0.01$ & 0.35 & 0.47 & 0.33 & 0.17 & 5.1 \\
Valid experience & $1.00 \pm 0.00$ & 0.02 & 0.00 & 0.32 & 0.00 & 3.1 \\
Out-of-date experience & $0.23 \pm 0.01$ & 0.01 & 1.00 & 0.30 & 0.28 & 4.8 \\
\midrule
Out-of-date, instruction to read the docs & $0.25 \pm 0.01$ & 1.00 & 0.99 & 0.32 & 0.28 & 5.1 \\
Out-of-date, instruction to verify & $0.69 \pm 0.01$ & 0.07 & 1.00 & 1.00 & 0.90 & 10.6 \\
% End of tables/ambig_main.tex
\bottomrule
\end{tabular}
\caption{Incomplete documentation: the tool definition and the documentation give the range of each value but not its
meaning, and the earlier request says what the values mean on its version. Success $\pm$ one standard error; process
measures as in Table~\ref{tab:main}; second write: share of runs with more than one write. The two instructions are
appended to the final task message (Section~\ref{sec:remedies}). Four changes, six models.}
\label{tab:undoc}
\end{table*}

\paragraph{Documentation.} In both versions the tool definitions and the documentation give only the type or range.
The priority tool documents ``\code{priority} is an integer from 1 to 4.'', the review-day tool ``\code{day} is an
integer from 0 to 6, one value for each day of the week.'', the review-month tool ``\code{month} is an integer, one
value for each month of the year.'', and the calendar documents \code{duration} as ``Length of the event (required)''.
The examples in the documentation are the same in both versions. The read tools return the priority as Highest, High,
Medium or Low, the review day and month by name, and an event with its start, end and duration in minutes. The month
change (version A: 1 is January; version B: 0 is January) and the duration change (minutes or seconds) are used only with
incomplete documentation; a value outside the range of the running version leaves the stored state unchanged.

\paragraph{Earlier request.} The earlier request ends with a sentence that says what the values mean on its version:
``(For this service, priority 1 is the highest and 4 is the lowest.)'' or ``\dots{} priority 4 is the highest and 1 is
the lowest.''; ``(For this service, review days are numbered 0 = Monday to 6 = Sunday.)'' or ``\dots{} 0 = Sunday to
6 = Saturday.''; ``(For this service, months are numbered 1 = January to 12 = December.)'' or ``\dots{} 0 = January to
11 = December.''; ``(For this service, the event length is given in minutes.)'' or ``\dots{} in seconds.''. The final
request has no such sentence.

\paragraph{Instructions.} The instruction to read the documentation appended to the final task message is the one in
Appendix~\ref{app:followup}; the instruction to verify has the wording of the system-prompt instruction
(Appendix~\ref{app:prompts}) and is appended to the final task message.

\paragraph{Runs and exclusions.} Twelve task instances per change, direction and arm; task titles are drawn from 24
titles. Of 2,880 episodes, 29 are excluded because the earlier task never succeeded in ten attempts: Inkling 27, all on
the weekday version in which 0 is Monday, and DeepSeek-V4-Flash 2, on the month version in which 0 is January. There
were no failed model requests. Without experience, success is 49\% and 71\% on the two priority versions, 43\% and 86\% on the two
weekday versions, 100\% and 40\% on the two month versions (1 or 0 is January), and 100\% and 17\% on the two duration
versions (minutes or seconds). Table~\ref{tab:undoc} gives the process measures, and Tables~\ref{tab:undoc-change} and~\ref{tab:undoc-model}
give the results by change and by model.

\begin{table*}[t]
\centering
\small
\setlength{\tabcolsep}{4pt}
\begin{tabular}{lccccc}
\toprule
 & & & & \multicolumn{2}{c}{Out-of-date, instruction to} \\
\cmidrule(lr){5-6}
Change & No experience & Valid & Out-of-date & read the docs & verify \\
\midrule
% Source inlined from tables/ambig_change.tex
Priority direction & $0.60 \pm 0.02$ & $1.00 \pm 0.00$ & $0.19 \pm 0.01$ & $0.22 \pm 0.02$ & $0.62 \pm 0.03$ \\
Weekday numbering & $0.65 \pm 0.02$ & $1.00 \pm 0.00$ & $0.27 \pm 0.02$ & $0.34 \pm 0.03$ & $0.78 \pm 0.03$ \\
Month numbering & $0.70 \pm 0.01$ & $1.00 \pm 0.00$ & $0.31 \pm 0.02$ & $0.30 \pm 0.02$ & $0.81 \pm 0.03$ \\
Duration unit & $0.58 \pm 0.00$ & $0.99 \pm 0.01$ & $0.15 \pm 0.01$ & $0.14 \pm 0.01$ & $0.56 \pm 0.03$ \\
% End of tables/ambig_change.tex
\bottomrule
\end{tabular}
\caption{Incomplete documentation by change: success $\pm$ one standard error.}
\label{tab:undoc-change}
\end{table*}

\begin{table}[!htbp]
\centering
\footnotesize
\setlength{\tabcolsep}{3pt}
\begin{adjustbox}{max width=\linewidth}
\begin{tabular}{lcccccc}
\toprule
 & & & & \multicolumn{2}{c}{+ instruction} & Read-back \\
\cmidrule(lr){5-6}
Model & None & Valid & Out-of-date & docs & verify & (out-of-date) \\
\midrule
% Source inlined from tables/ambig_model.tex
GLM & 0.85 & 1.00 & 0.40 & 0.46 & 0.82 & 0.55 \\
Muse & 0.43 & 1.00 & 0.02 & 0.01 & 0.21 & 0.20 \\
DeepSeek & 0.49 & 1.00 & 0.00 & 0.04 & 0.74 & 0.00 \\
MiniMax & 0.52 & 0.98 & 0.01 & 0.01 & 0.75 & 0.01 \\
Qwen3.8 & 1.00 & 1.00 & 0.93 & 0.93 & 0.96 & 1.00 \\
Inkling & 0.50 & 1.00 & 0.00 & 0.03 & 0.67 & 0.00 \\
% End of tables/ambig_model.tex
\bottomrule
\end{tabular}
\end{adjustbox}
\caption{Incomplete documentation by model: success, and the read-back rate with out-of-date experience.}
\label{tab:undoc-model}
\end{table}

\paragraph{Cost of the instruction with valid experience.} With complete documentation and the five typed changes, we
also appended the instruction to read the documentation to the final task message when the experience was valid (360
episodes, the same task instances as Table~\ref{tab:context}). Success stayed at 100\% (100\% without the instruction);
the share of runs that search the documentation before the first write rose from 1\% to 100\%, model calls per task
from 2.92 to 3.55, input tokens from 3,839 to 5,361 and output tokens from 224 to 346. Six episodes were excluded
because the earlier task never succeeded.

\section{Calibration and Exclusions}
\label{app:calibration}

Before the run, a calibration run with all six models, one instance per cell and one direction (1,062 episodes) checked
four rules fixed in advance: C1, failed model requests in more than 3\% of a model's episodes; C2, success with valid experience below
70\% in any model and condition; C3, more than 10\% of valid-experience runs reaching the step limit; C4, the
earlier task never succeeding in more than 10\% of a cell. Only C4 was triggered, for Inkling in two of the additions to
the final task, all on the weekday change; as the rule specified, the number of attempts for the earlier task was raised
from seven to ten. The calibration episodes are not used.

Across all conditions, 136 of 12,624 episodes are excluded: 9 because a model request failed (MiniMax-M3 8, Muse-Glimmer-30B 1) and
127 because the earlier task never succeeded in ten attempts (Inkling 76, DeepSeek-V4-Flash 36, MiniMax-M3 15). Of the
latter, 124 are on the weekday change, two on the month change and one on another change. Of all earlier tasks in the
included episodes with complete documentation, 97\% succeeded at the first attempt. Valid-experience runs reached the
step limit in fewer than 1\% of cases.

\end{document}